\documentclass{article}

\usepackage[creativeai, final]{neurips_2025}

\usepackage[utf8]{inputenc} %
\usepackage[T1]{fontenc}    %
\usepackage{hyperref}       %
\usepackage{url}            %
\usepackage{booktabs}       %
\usepackage{amsfonts}       %
\usepackage{nicefrac}       %
\usepackage{microtype}      %

\usepackage{caption}
\usepackage{capt-of}
\usepackage{graphicx}
\usepackage{amsmath}
\usepackage{xcolor}         %
\usepackage[capitalize]{cleveref}
\usepackage{wrapfig}
\crefname{section}{Sec.}{Secs.}
\Crefname{section}{Section}{Sections}
\Crefname{table}{Table}{Tables}
\crefname{table}{Tab.}{Tabs.}

\newcommand{\paragraphcustom}[1]{\noindent\textbf{#1}}

\title{Opt-In Art: Learning Art Styles Only from Few Examples}

\author{%
  Hui Ren$^{1,2,}$\thanks{Equal contribution.}\,\,\,$^{,}$\thanks{Work done while visiting MIT CSAIL.} \quad
  Joanna Materzy\'nska\footnotemark[1] \quad
  Rohit Gandikota$^{3}$ \quad
  Giannis Daras$^{2}$ \quad \\
  \textbf{David Bau}$^{3}$ \quad
  \textbf{Antonio Torralba}$^{2}$ \\
  \\
  $^{1}$UIUC \quad $^{2}$MIT \quad $^{3}$Northeastern University \\
  \\
  \texttt{\{huiren2@illinois.edu, jomat@mit.edu, gandikota.ro@northeastern.edu,} \\
  \texttt{gdaras@mit.edu, d.bau@northeastern.edu, torralba@mit.edu\}} \\
  \\ \vspace{3pt}
\href{https://joaanna.github.io/art-free-diffusion}{\textbf{\textcolor{purple}{\texttt{Project Webpage}}}}
}

\begin{document}
\maketitle
\vspace{-1cm}
\begin{center}
    \centering

    \includegraphics[width=\linewidth]{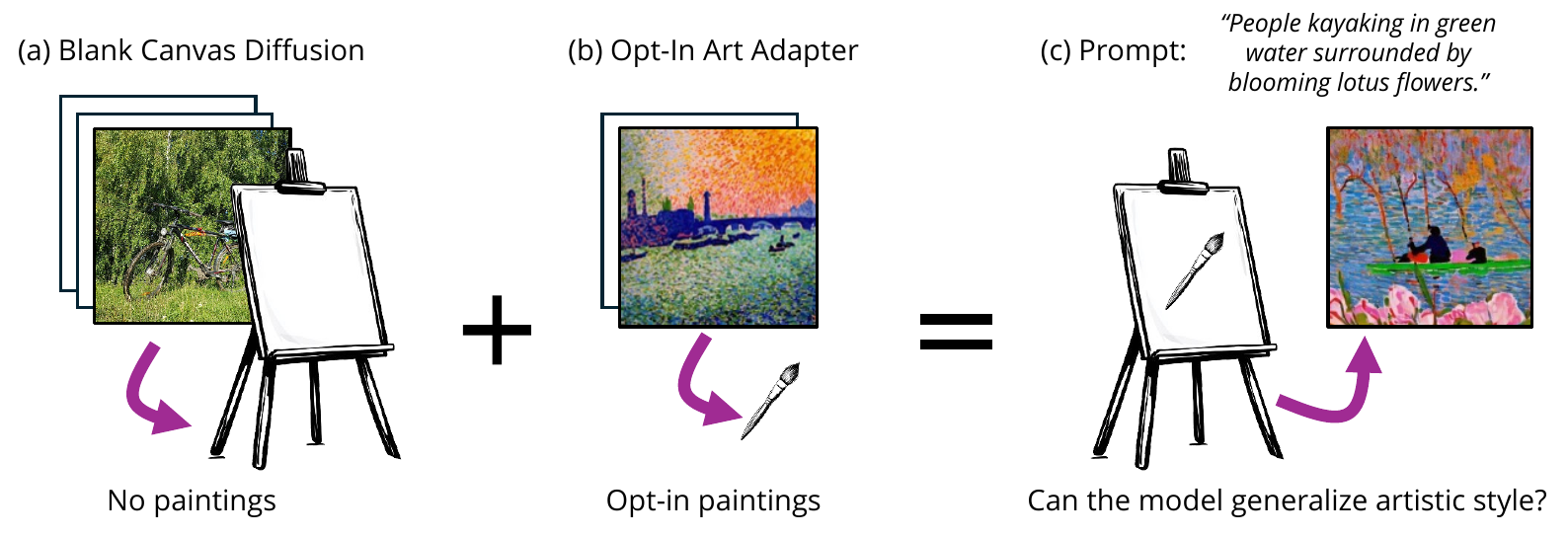}

    \captionof{figure}{(a) We introduce Blank Canvas Diffusion, a carefully curated text-to-image model trained only on photographs, serving as the pretraining foundation for our model. Our study explores whether a model with no prior exposure to paintings can learn artistic styles using (b) a LoRA Art Adapter trained on a small opt-in sample of an artist's work. (c) We find that it is possible to adapt a model that is trained without paintings to generalize an artistic style, given only few examples.}
    \label{fig:teaser}
\end{center}

\vspace{-0.3cm}

\begin{abstract}
We explore whether pre-training on datasets with paintings is necessary for a model to learn an artistic style with only a few examples. To investigate this, we train a text-to-image model exclusively on photographs, without access to any painting-related content. 
We show that it is possible to adapt a model that is trained without paintings to an artistic style, given only few examples. User studies and automatic evaluations confirm that our model (post-adaptation) performs on par with state-of-the-art models trained on massive datasets that contain artistic content like paintings, drawings or illustrations.
Finally, using data attribution techniques, we analyze how both artistic and non-artistic datasets contribute to generating artistic-style images. Surprisingly, our findings suggest that high-quality artistic outputs can be achieved without prior exposure to artistic data, indicating that artistic style generation can occur in a controlled, opt-in manner using only a limited, carefully selected set of training examples.

\end{abstract}

\vspace{-0.1in}
\section{Introduction}
\label{sec:introduction}
\vspace{-0.05in}

In this work, we aim to disentangle a generative model’s ability to create artistic imagery from its reliance on prior exposure to human-created paintings, drawings, or illustrations.
The recent success of generative models, particularly denoising diffusion models \citep{ho2020denoisingdiffusionprobabilisticmodels}, introduces yet another tool for artistic expression. Unlike traditional artistic tools such as cameras or animation software, which are rarely questioned in terms of their creative agency, generative modeling challenges conventional notions of authorship and artistic ownership. The ability of these models to generate images in the style of specific artists \citep{gandikota2023erasing} raises pressing questions: Who holds the rights to AI-generated images—the original artist, the model’s developer, or the user providing the prompt \citep{epstein2023art, Whiddington2024}? Furthermore, how much of the artistic output is a genuine act of creation versus a statistical mimicry of preexisting styles \citep{epstein2023art, somepalli2023diffusion, somepalli2024measuring}?
Unlike traditional artistic tools, generative models learn from large datasets of human-created art, enabling them to replicate styles with high fidelity~\citep{Simonite2022}. This blurs the line between inspiration and imitation, raising concerns about creative displacement and style homogenization~\citep{epstein2023art}, and highlights the need to examine their dependence on existing artworks.

Existing models are trained on massive datasets like LAION-5B~\citep{schuhmann2022laion}, but it remains unclear whether their artistic ability stems from direct exposure to art or general visual learning. To probe this, we train \textbf{Blank Canvas Diffusion} exclusively on photographs, using the \textbf{Blank Canvas Dataset}, a rigorously filtered collection free of paintings, drawings, and illustrations.  
We then apply the \textbf{Art Style Adapter}, a fine-tuning framework that enables the model to learn artistic styles from just a few examples in a controlled, \textbf{Opt-In Art} setup. We evaluate via painting similarity, crowd-sourced style judgments, and data attribution. Despite no prior exposure to art, our model effectively replicates artistic styles, performing comparably to models trained on art-rich datasets.

These findings suggest that generative models need not rely on large-scale artistic datasets to learn an artistic style. Even without prior exposure to paintings or illustrations, a model can learn to generate art through minimal fine-tuning on a few examples. This offers a pathway to creative AI systems that respect artistic consent. However, our results also complicate the opt-in model debate: if only a few reference images suffice for style replication, restricting training data may be insufficient to protect artistic styles. This raises concerns about the limits of current copyright protections~\citep{lu2024disguised}, as artists may still have little control over the use of their style. Future regulatory discussions may need to go beyond data restrictions to address attribution, consent, and fair use.

\vspace{-0.25cm}
\section{Related work}
\vspace{-0.25cm}

\begin{table*}[htbp]
    \centering
    \resizebox{\textwidth}{!}{%
        \begin{tabular}{l|cccccccccccccccccccc}
            \toprule
            & \#sample & paintings & stamp & sculptures & digital art & logo & artwork & sketch & advertisement & drawing & illustration & installation art & mosaic art & tapestry & baroque art & art noveau & pop art & total \\
            \midrule
            SA-1B      & 10,000 & 36 & 71 & 120 & 14 & 36 & 0 & 0 & 2 & 8 & 4 & 12 & 1 & 3 & 6 & 1 & 2 & 315 (~3.15\%) \\
                        Blank Canvas Dataset & 10,000 & 0  & 0  & 38  & 1  & 12  & 0 & 0 & 2 & 3 & 0 & 12 & 2 & 1 & 0 & 0 & 0 & 71 (~0.71\%) \\

            \bottomrule
        \end{tabular}
    }
    \caption{Statistics of artistic images found during manual inspection of the SA-1B and Blank Canvas datasets before and after the art filter.}
    \label{fig:filtering_table_numbers}
    \vspace{-0.2in}

\end{table*}

\begin{wrapfigure}{r}{0.45\linewidth}
    \centering
    \vspace{-0.8cm} %
    \includegraphics[width=\linewidth]{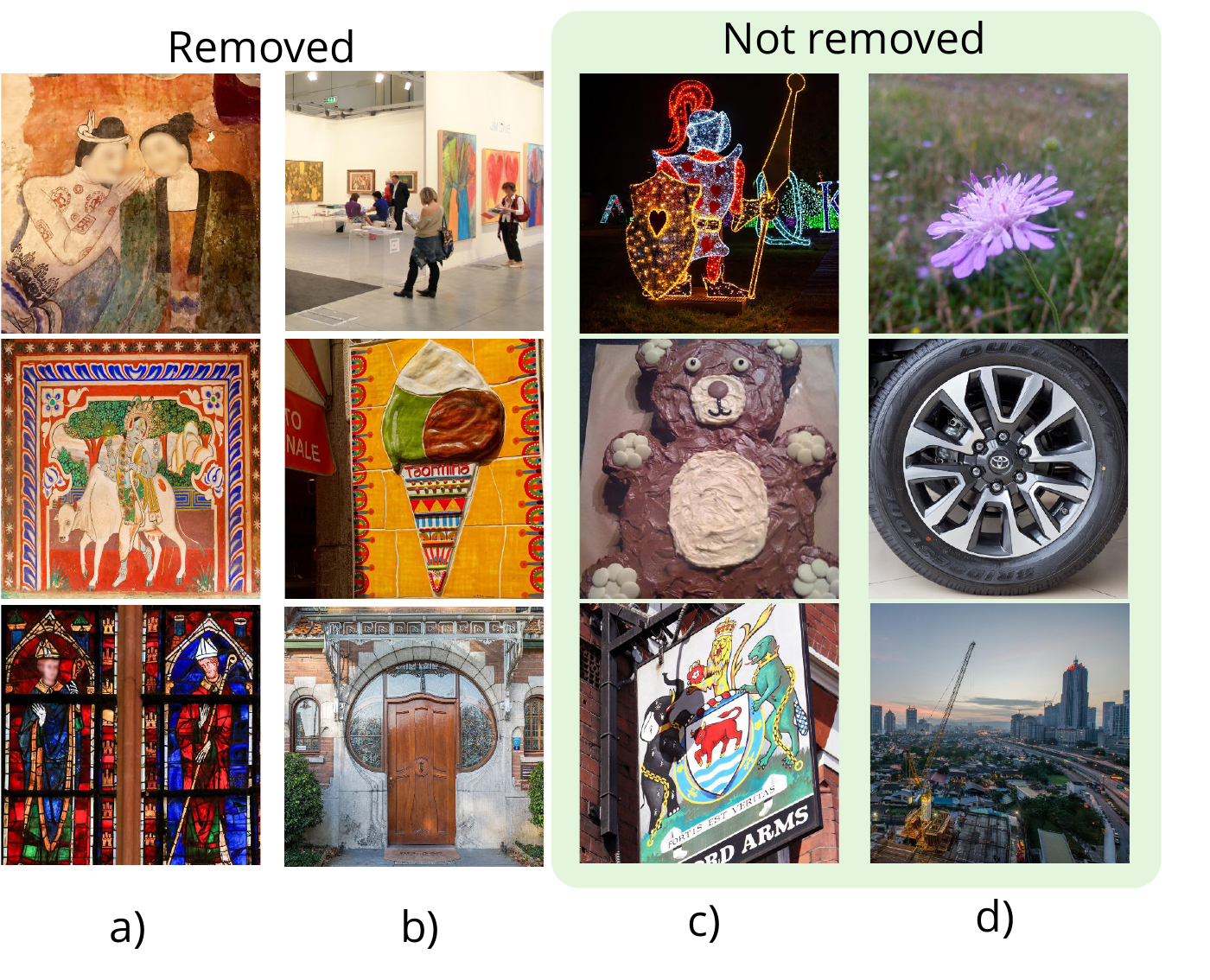}
    \vspace{-0.5cm}
    \caption{Examples of images included and excluded from the Blank Canvas dataset.
    The dataset is curated to remove paintings as well as artistic categories related to paintings, such as drawings and fine art. Examples of images that are close to the removal threshold are shown in b) and c).}
    \label{fig:prior_knowledge}
    \vspace{-0.8cm}
\end{wrapfigure}

Diffusion models memorize more training data compared to GANs and VAEs, raising ethical concerns~\citep{somepalli2023diffusion, somepalli2024measuring, carlini2023extracting}. Several mitigation strategies have been proposed: concept erasure methods \citep{gandikota2023erasing, gandikota2023unified, kumari2023ablating}, opt-out initiatives \citep{spawningAI}, watermarking approaches \citep{zhao2023recipe, min2024watermark}, and training with corrupted data to avoid exact replication \citep{somepalli2024measuring}.
Perhaps the most straightforward approach is curating copyright-free training datasets. 
Gokaslan et al.\citep{gokaslan2024commoncanvas} train a model on Creative Commons images, but still include artistic content.
We extend this by training with minimal exposure to paintings, drawings, and artistic media, creating a diffusion model largely devoid of non-photographic art (Fig. \ref{fig:prior_knowledge}). 
We investigate how few artistic images are required post-training to learn specific styles.
To adapt our art-agnostic model, we use our Art Style Adapter based on LoRA fine-tuning~\citep{hu2021lora} and Textual Inversion~\citep{gal2022textual}. Prior work shows methods like Textual Inversion and Dreambooth~\citep{ruiz2023dreamboothfinetuningtexttoimage} can personalize diffusion models using few user images. Our novel finding is that these adaptation methods work even when the base model has never been trained on the target image type.

\vspace{-0.5cm}
\section{Blank Canvas Diffusion}
\paragraphcustom{Blank Canvas dataset.}  
Most diffusion models are trained on large-scale datasets that include paintings, drawings, and digital art. Our goal is to create a large-scale dataset composed exclusively of photographs. To this end, we curate the \textbf{Blank Canvas dataset} from the SAM-LLaVA-Captions10M dataset~\citep{chen2023pixartalpha}, derived from SA-1B~\citep{kirillov2023segment}, a dataset of natural, camera-captured images with captions generated by a vision-language model (LLaVA).  

Despite its photographic intent, the dataset contains incidental artwork e.g., photos of tapestries, logos, or sculptures. We therefore design a two-stage filtering pipeline to minimize art content. We exclude images via caption-based keyword filtering, then we compute CLIP-based cosine similarity scores to art-related terms and remove high-scoring images. Thresholds are determined by manual inspection. After filtering, the dataset retains 9.11M image-text pairs (4.7\% removed via text, 16.7\% via image filtering). Manual inspection confirms a reduction in art content: in SA-1B, 315 of 10k images originally contained artwork, versus 72 post-filtering. Similar evaluation on COCO shows a drop from 1.06\% to 0.12\%. 
We also estimate that 32.9\% of images in LAION-Aesthetic v2 5+ (used to train Stable Diffusion 1.4) contain artwork, suggesting over 191M art images in its training set.

Our \textbf{Blank Canvas Diffusion} model uses a latent diffusion architecture~\citep{rombach2022highresolution} with three modules: a VAE, a U-Net, and a Text Encoder. To ensure no exposure to artistic content, we train both the VAE and U-Net from scratch on our filtered dataset. Unlike prior models that use CLIP~\citep{radford2021learning}, which may encode visual artistic concepts, we use a BERT-based language, only encoder~\citep{devlin2019bert}. While BERT may contain textual knowledge of art, it has no visual grounding, ensuring that the model is free from any pixel-level artistic priors.

\begin{figure*}[htbp]
    \centering
    \includegraphics[width=1\textwidth]{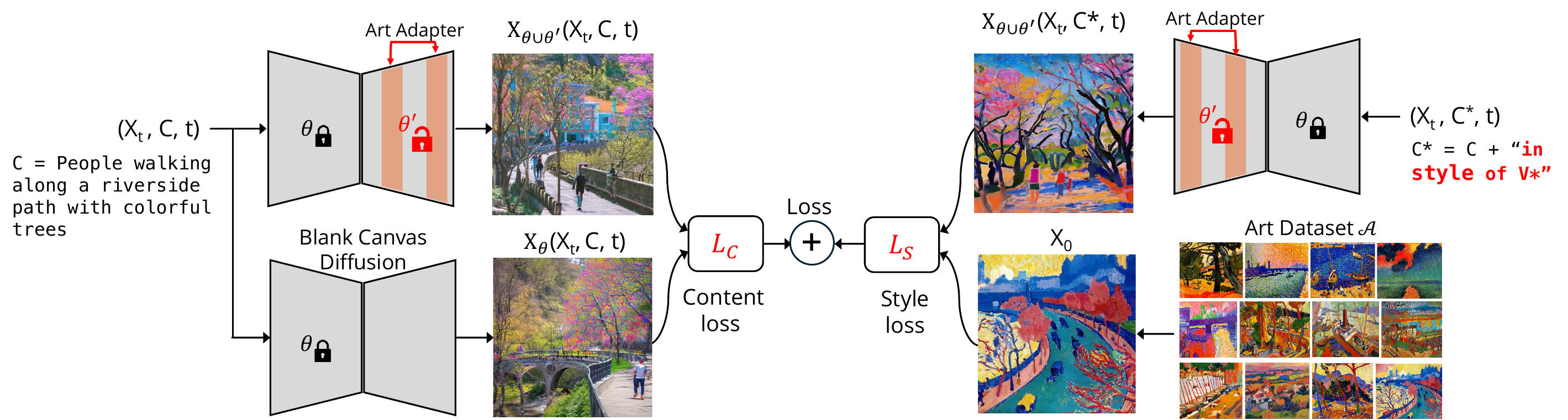}
    \caption{The generated image should match the style of a small exemplar dataset when prompted with a caption \( C^* \), which includes a style prefix \( V^* \). For example, if \( C^* = \textit{People walking along a riverside path with colorful trees in the style of } V^* \), the image should reflect both the scene (content) and the specified artistic style. Content loss ensures that the visual elements of the prompt \( C = \textit{People walking along a riverside path with colorful trees} \) are accurate, while style loss maintains the style associated with \( V^* \).}

    \label{fig:art_inject}
\end{figure*}
\vspace{-0.2cm}

\section{Opt In: Artistic Style Adapter}
\label{sec:art_inject}

To opt-in to an art style, we train a LoRA Art-Style Adapter by collecting a few examples of paintings in that style $X_0 \in \mathcal{A}$ and caption the content of the artwork. Captioning can be done automatically or manually. To connect the newly learned style information with specific tokens in the prompt, we append ``in the style of V* art" to the content prompt. We use $C^*$ for the final prompt (after our addition). To enable the model to learn this new artistic style, we fine-tune the U-Net module using LoRA
\citep{hu2021lora}. For a given target artistic image, we define the following loss:
\begin{equation}
    \mathcal{L}_{\text{S}}(\theta') = \left\| \epsilon_{\theta \cup \theta ^\prime} (X_t, C^*, t) - \epsilon |\right\|^2,
\end{equation}
where $\epsilon_{\theta \cup\theta ^\prime} $ is the U-Net module with the LoRA updating weights, $t$ is the denoising time step, $X_t$ is the input image at time $t$, and $\epsilon$ is target noise. We refer to this loss as style loss, as it helps the model implicitly learn the artistic style and link it to the style modification in the prompt. The content loss is defined as follows:
\begin{equation}
    \mathcal{L}_{\text{C}}(\theta') = \left\| \epsilon_{\theta \cup \theta ^\prime} (X_t, C, t) - \epsilon_{\theta} (X_t, C, t) |\right\|^2.
\label{eq:prior_loss}
\end{equation}
This loss helps maintain the prompt's content even when the style identifier is omitted from the text.
Our final loss is $\mathcal{L} = \mathcal{L}_{\text{S}}+ w \cdot \mathcal{L}_{\text{C}}$, where $w$ is the hyper-parameter for the content loss. We combine style and content losses to prevent the model from overfitting to artistic features, allowing it to learn style as a distinct component separate from content. This approach encourages the model to capture the underlying style patterns without embedding them too deeply into the content, enabling it to generate natural images when no specific style is specified. By disentangling style from content in this way, the model learns to apply styles more flexibly while preserving the core content. 
At inference, we control style by adjusting when art information is introduced. Injecting style earlier makes the image more stylized, later injection preserves natural details with subtle artistic elements.

\section{Experiments}
\label{sec:experiments}

\paragraphcustom{Blank Canvas Diffusion} Our Blank Canvas Diffusion follows Stable Diffusion v1.4 architecture \citep{rombach2022highresolution}. We train VAE from scratch, then U-Net on Blank Canvas dataset with frozen VAE, using pre-trained BERT \citep{devlin2019bert}.
We compare against CommonCanvas-SC \citep{gokaslan2024commoncanvas} (30M CC images) and Stable Diffusion v1-4 (600M images), in terms of data, compute time, image quality (FID score) and text and image alignment (CLIP score). Compared to existing models, Blank Canvas Diffusion is trained with significantly fewer resources. It uses only 9 mln image-text pairs and 11,432 A100 GPU hours. In contrast, CommonCanvas-SC is trained on 30 mln images with 73,800 A100 hours, and Stable Diffusion v1.4 (SD1-4) is trained on 600M images using 150,000 A100 hours.
Despite this large gap in data and compute, our model achieves a CLIP score of 0.26 and FID of 12.12 on the Blank Canvas test set, competitive with CommonCanvas-SC (0.27 / 13.66) and SD1-4 (0.28 / 17.74). On the COCO-2017 test set, where our model faces a domain mismatch, it obtains 0.23 CLIP and 23.60 FID, while CommonCanvas-SC and SD1-4 perform better (0.27 / 8.23 and 0.27 / 12.54, respectively). These results underscore the competitiveness of our model despite operating at a fraction of the data and compute budget.

\paragraphcustom{Art Style Adaptation} We adapt styles using our Art Adapter with content loss weight $w=50$. From WikiArt, we select 17 artists with distinct styles and manually curate 9–50 paintings per artist (avg. 21.88), ensuring stylistic consistency in color, brushwork, and content. We evaluate style similarity using the Contrastive Style Descriptor (CSD)~\citep{somepalli2024measuring}. For each generated image, we compute its mean CSD score against the corresponding artist set. Content preservation is assessed via cosine similarity in ViT-based content features ($\rm ViT_c$), and CLIP score measures text-image alignment.

\begin{figure*}[htbp]
    \centering
\includegraphics[width=\linewidth]{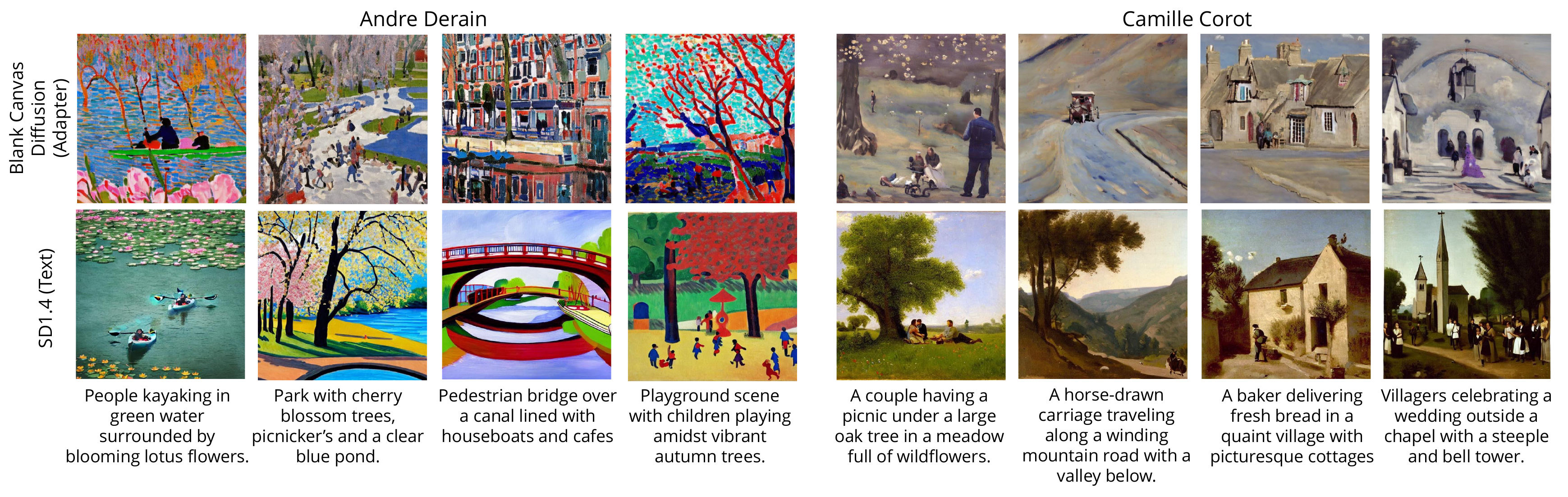}
    \caption{Comparison of Blank Canvas Diffusion art generation (top row) with Stable Diffusion 1.4 generated images (bottom row).}
    \label{fig:art_gen}
\end{figure*}

We sample 500 prompts and images from the LAION Pop dataset~\citep{schuhmann2023laion}, evaluating across 17 artist style sets with results averaged per style. To validate performance, we conduct a user study via Amazon Mechanical Turk, comparing outputs from our Art Adapter and baseline methods on two tasks: Image Stylization and Artistic Style Generation.  Participants are shown three reference images from a given artist and asked to select which of two outputs best matches the style. We also include comparisons with real paintings. To mitigate bias toward photographic realism~\citep{wang2023computational}, we include reliability checks, yielding 2,242 responses from 42 users after filtering.

\begin{figure}[htbp]
    \centering
    \includegraphics[width=1\linewidth]{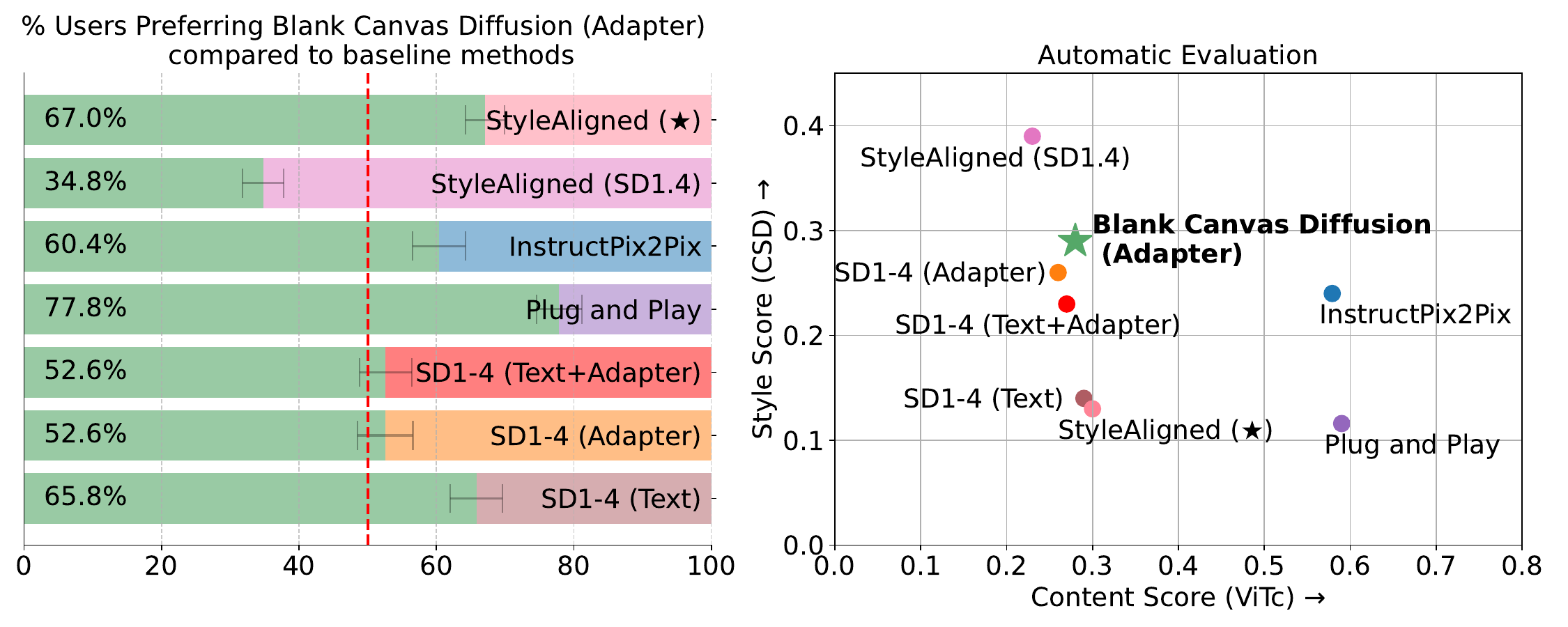}
    \caption{(Left) Results of the Perceptual User Study; Blank Canvas Diffusion with Adapter method (green bar) is preferred over image editing baselines, is on par with Adapter using an SD1.4 backbone and is favored less than StyleAligned (SD1.4). The margin of preference is narrow between the baselines. (Right) Quantitative evaluation; Blank Canvas Diffusion with the Art Adapter achieves a good trade-off between the style and content.}
    \label{fig:img_stylizationq}
    \vspace{-0.15in}
\end{figure}

\paragraphcustom{Image stylization.}  
We evaluate our method on image stylization using the LAION Pop dataset, aiming to transfer artistic style while preserving image content. Baselines include: \textbf{SD1.4 (Adapter)} using a learned style adapter and new token, \textbf{SD1.4 (Text)} using only the artist's name, and \textbf{SD1.4 (Adapter + Text)} combining both. For all SD1.4 variants, LoRAs are applied across all blocks for optimal performance. Stylization is performed via DDIM inversion to step 800, followed by denoising with a modified prompt and adapter. We also compare to Plug and Play~\citep{Tumanyan_2023_CVPR}, InstructPix2Pix~\citep{brooks2023instructpix2pix}, and StyleAligned~\citep{hertz2024style}, as well as CycleGAN~\citep{zhu2017unpaired} for Monet and Van Gogh. Qualitative results for imitating Van Gogh's style are shown in SM.\ref{sup:qual_imgstyl}.
Our method matches or outperforms all baselines in both human and automatic evaluations. In user studies (Fig.~\ref{fig:img_stylizationq}), participants consistently preferred our outputs over Plug and Play, InstructPix2Pix, and StyleAligned on Blank Canvas Diffusion. While StyleAligned on SD1.4 showed slightly higher preference, it relies on extensive art-pretraining, unlike our model. Automatic metrics support these findings: our method balances style (CSD) and content (ViT$_c$) fidelity, whereas baselines often favor one at the expense of the other (Tab. \ref{fig:img_stylizationq} (Right)).

\begin{figure}[htbp]
    \centering
\includegraphics[width=1\linewidth]{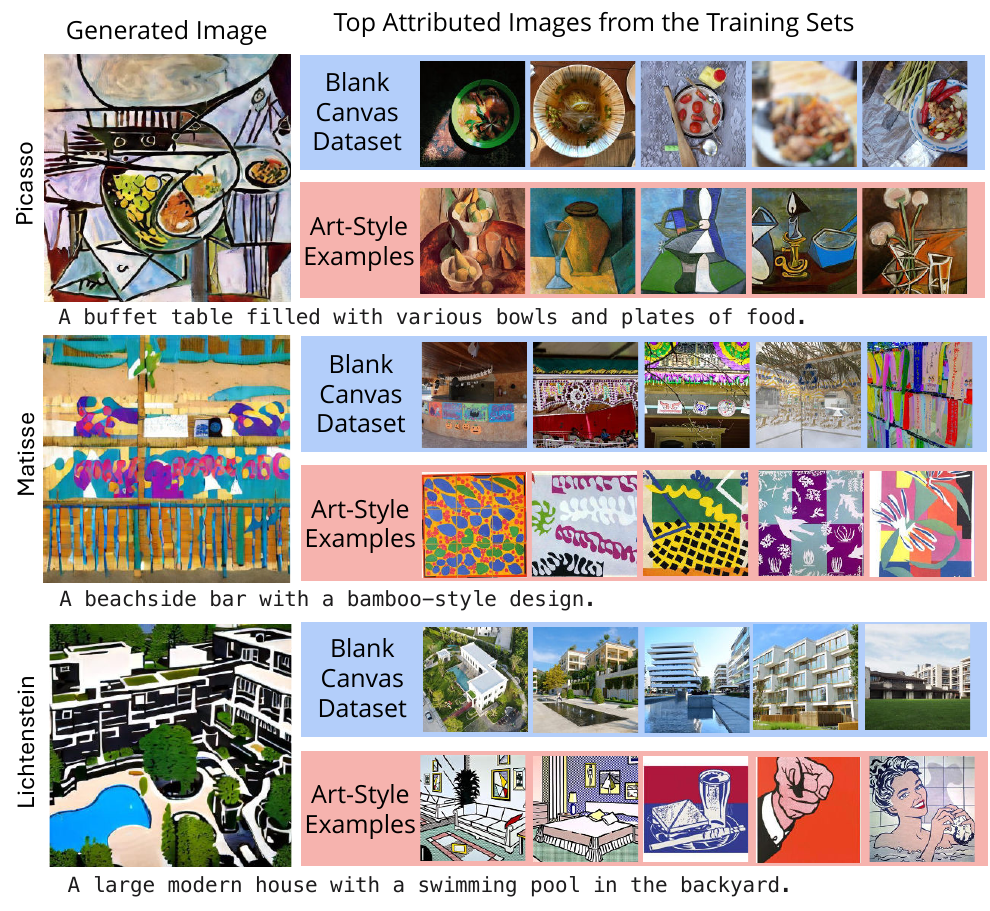}
\vspace{-0.1in}
    \caption{Data attribution experiments on stylized images reveals that while the generated images reflect the distinct artistic styles of each artist, the training images that contributed the most came from both the Blank Canvas dataset and the Art-Style examples.}
\vspace{-0.6cm}
    \label{fig:attr}
\end{figure}

\paragraphcustom{Artistic Style Generation} 
We address the task of Artistic Style Generation, focusing on creating images in a specific artistic style. Stable Diffusion, known for its ability to replicate styles by simply prompting with artist names, serves as a baseline due to its extensive training on artworks \cite{heikkila_2022}. Additionally, we compare with transferring a style into generated images with the StyleAligned method on both Stable Diffusion and Blank Canvas Diffusion backbones.
Qualitative examples presented in Fig. \ref{fig:art_gen}. Blank Canvas Diffusion (Adapter) outperforms SD1-4 (Text) in style (CSD 0.34 vs. 0.22) but slightly lags in content preservation (0.21 vs. 0.26). StyleAligned scores 0.47 on Stable Diffusion but drops to 0.22 on Blank Canvas Diffusion, highlighting the need for an artistic backbone.
Our perceptual study confirms these trends: 76.2\% of participants preferred our method over SD1-4 (Text) for style. Against StyleAligned with SD1-4, our model was chosen 31.5\% of the time, reinforcing SD1-4’s strong artistic capacity from extensive pretraining.
When asked to identify images closest to real artworks, participants selected our method 17.5\% of the time and SD1.4 (Text) 11.1\%, suggesting our approach better mimics authentic artistic styles.

\paragraphcustom{Data Attribution}
We find that our Art Adapter generalizes well from a small art-style training set, generating novel images consistent with the target style. To understand which training images influenced these generations—and to check whether art content may have inadvertently remained in the training set—we apply the data attribution method from~\citet{wang2023evaluating}. Results are shown in Fig.~\ref{fig:attr}, where we retrieve the top five attributed images from both the Blank Canvas Dataset and the Art-Style examples for each output. While stylistic features often dominate, real-world influences from the Blank Canvas Dataset remain significant. In the Picasso-style example, the output reflects cubist elements, and attribution reveals all top five matches from the Art-Style dataset. In contrast, for Matisse and Lichtenstein styles, the top attributed images come from the Blank Canvas Dataset. The Matisse-style image shows vivid colors and organic shapes, yet attribution reveals underlying real-world scenes. Similarly, in the Lichtenstein example, the bold comic style overlays content clearly traced back to non-artistic pretraining images.

\section{Discussion and Limitations}
\vspace{-0.1in}

Blank Canvas Diffusion explores whether generative models can learn to imitate artistic style with minimal exposure to traditional artworks. Using a simple Art Style Adapter, we show that models trained on largely art-free data can still generate stylistically rich images. Attribution analysis suggests that patterns in natural imagery may already carry the foundations of visual style, offering insight into how generative models internalize and repurpose structure in the world around them.
While we took care to filter explicit artworks, we acknowledge that photography itself is not free of aesthetic intent. Choices in composition, lighting, and framing can carry stylistic influence \citep{hertzmann2022choices}, and some images in our dataset may reflect these qualities. Although not central to our findings, this nuance highlights the difficulty of fully disentangling artistic influence from visual data.
Our results suggest that generative models may not require direct exposure to traditional art to develop artistic capabilities, complicating assumptions about what kinds of data drive creativity and raising broader questions about influence, authorship, and control in generative systems.

\section{Acknowledgments}\label{sec:ack}
We thank Alan Kenny for his thoughtful discussions and for graciously allowing the use of his artwork in this research. We also appreciate Yael Vinker and Tamar Rott Shaham for their valuable feedback on our paper. JM is grateful for support from the ONR MURI grant (\#033697-00007), IBM (\#027397-00163), and Hyundai Motor Company R\&D (\#034197-00002). RH and DB are supported by Open Philanthropy and NSF grant \#2403304.

{
    \small
    \bibliographystyle{ieeenat_fullname}
    \bibliography{main}
}

\clearpage
\appendix

\section*{Supplementary Material}
\addcontentsline{toc}{section}{Supplementary Materials} 

\section{Diffusion Model Background}
\label{sup:preliminary}

Diffusion models \cite{ho2020denoisingdiffusionprobabilisticmodels} represent a class of generative models capable of producing high-quality images by modeling data distributions through successive denoising steps. In diffusion modeling we define a forward process that incrementally introduces noise to the data distribution, transforming it into a Gaussian distribution over time. At any given time step, the relationship between the image and the noise can be expressed as:
\begin{equation}
    X_t = \sqrt{1-\beta_t} \cdot X_0 + \beta_t \cdot \epsilon,
\end{equation}
where $X_t$ represents the image at time step $t$, $X_0$ is the original image, $\epsilon$ denotes a Gaussian Random Variable with zero mean and unit variance, and $\beta_t$ is the standard deviation of the added noise. Diffusion models optimize for the score function, $\nabla \log p_t(\cdot)$, needed to reverse the corruption process. The score can be learned with supervised learning using the Denoising Score Matching objective defined as:
\begin{equation}
    \min _{\theta} \mathbb{E} \left[ \left\| \epsilon_{\theta} (X_t, C, t) - \epsilon \right\|^2 \right],
\end{equation}
where $\epsilon_{\theta}$ is the model and C is the condition, in our case, the text prompt. For computational reasons, the corruption process often occurs in low dimensional latent space~\cite{rombach2022highresolution}.

\section{Artwork Filtering Methodology}
\label{sup:filtering}
Our artwork filtering process operates on both image and caption levels to ensure comprehensive coverage. For image-level filtering, we define a set of concepts to be excluded:

\begin{quote}
painting, art, artwork, drawing, sketch, illustration, sculpture, stamp, advertisement, logo, installation art, printmaking art, digital art, conceptual art, mosaic art, tapestry, abstract art, realism art, surrealism art, impressionism art, expressionism art, cubism art, minimalism art, baroque art, rococo art, pop art, art nouveau, art deco, futurism art, dadaism art
\end{quote}

\begin{figure}[htbp]
    \centering
    \includegraphics[width=0.8\linewidth]{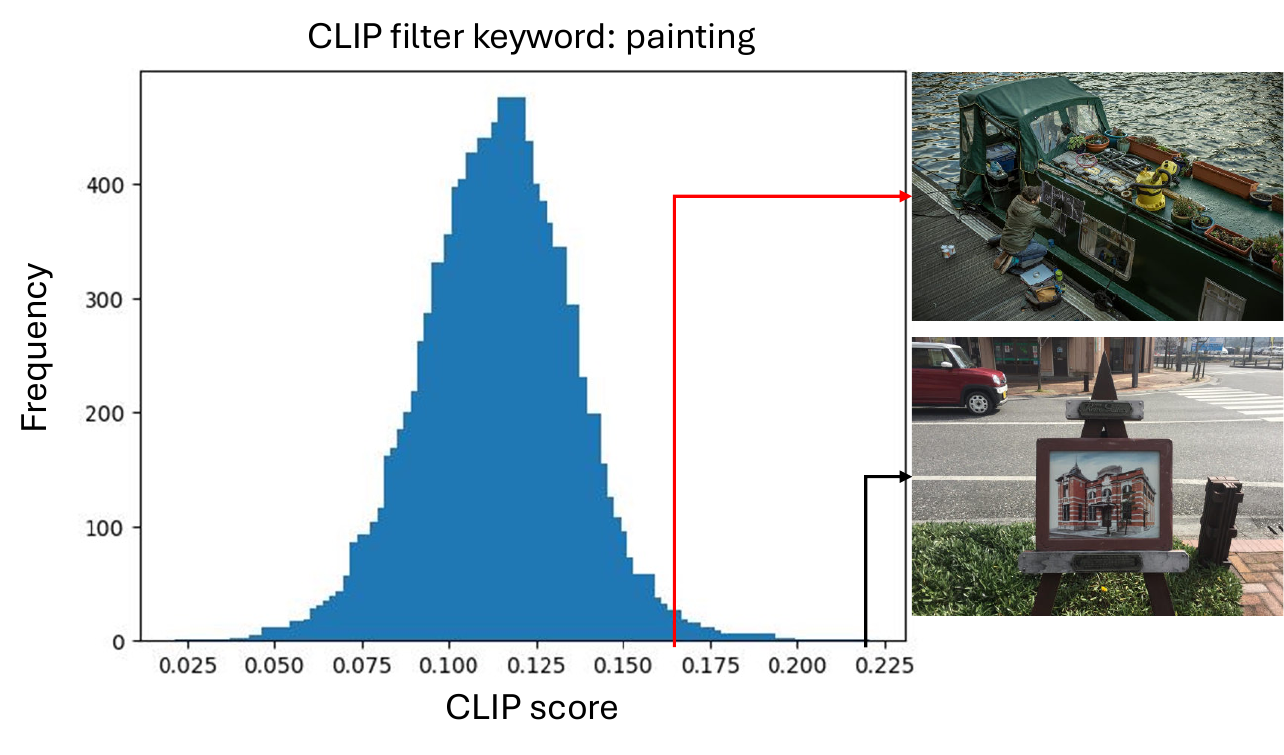}
    \caption{Histogram of the CLIP score of images with the word ``painting''. The distribution shown is from a subset of the SA-1B dataset. The red line represents the filtering threshold (17) we selected. Our strict threshold aims filters out all the art, even incidental art like a picture of a man painting.}
    \label{fig:filter_hist_painting}
\end{figure}

Fig. \ref{fig:filter_hist_painting} presents a histogram of CLIP scores for images associated with the word ``painting''. This distribution is derived from a subset of the SA-1B dataset, comprising 11,186 images (0.1\% of the complete SA-1B dataset).

For caption-level filtering, we exclude the following terms:

\begin{quote}
painting, paintings, art, artwork, drawings, sketch, sketches, illustration, illustrations, sculpture, sculptures, stamp, stamps, advertisement, advertisements, logo, logos, installation, printmaking, digital art, conceptual art, mosaic, tapestry, abstract, realism, surrealism, impressionism, expressionism, cubism, minimalism, baroque, rococo, pop art, art nouveau, art deco, futurism, dadaism
\end{quote}

\section{Manual Filtering}
\label{sup:maunal_filtering}

\begin{figure}[ht]
    \centering
    \includegraphics[width=\linewidth]{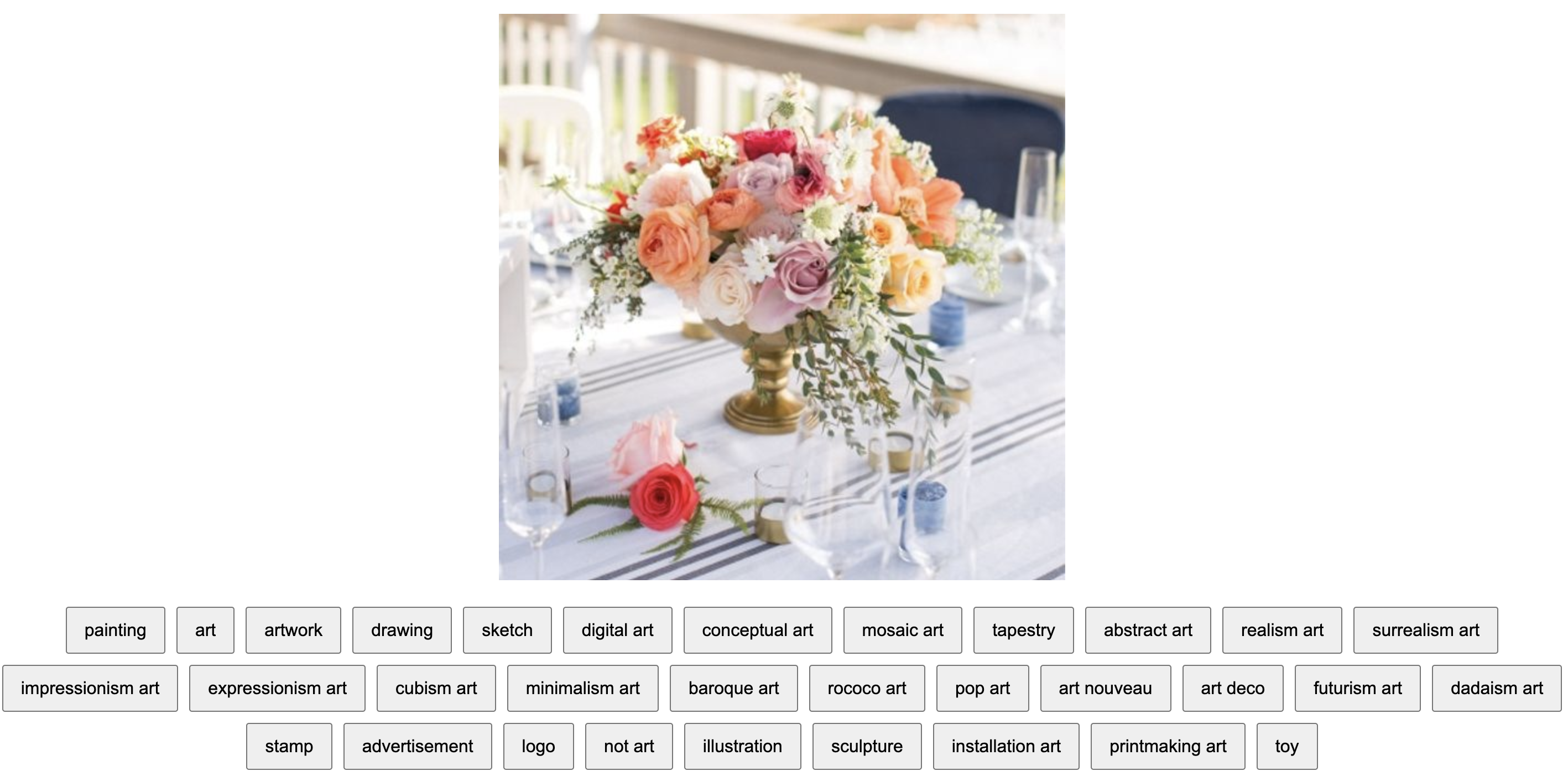}
   \caption{Manual inspection tool}
   \label{fig:sm-manual}
\end{figure}

\begin{figure}[ht]
    \centering
    \includegraphics[width=\linewidth]{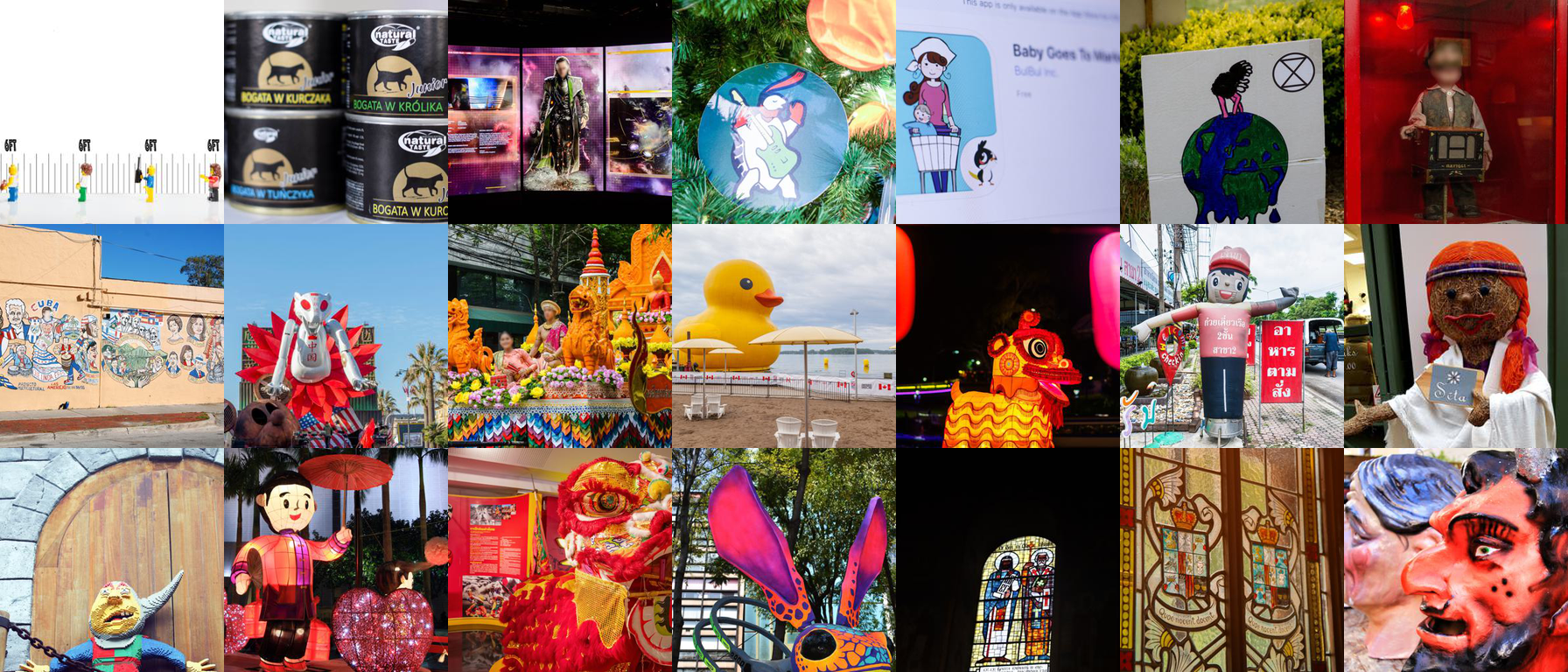}
   \caption{In a random sample of 10000 images (total: 71), we found categories; advertisement (2), digital art (1), drawing (3), installation art (12), logo (12), mosaic art (2), printmaking art (1), and sculpture (38), from left to right, excluding sculpture and logo.}
   \label{fig:sm-randomsample}
\end{figure}

To inspect the quality of our dataset, one of the authors manually inspected a random sample of 10k images. We did so by writing a simple javascript script that displayed individual images on a page, by clicking the category a result was written in a text file (see Fig. \ref{fig:sm-manual} --\ref{fig:sm-randomsample}).

\section{Qualitative Results of Blank Canvas Diffusion}
\label{sup:qual_artfree}

\begin{figure}[htbp]
    \centering
    \includegraphics[width=0.8\linewidth]{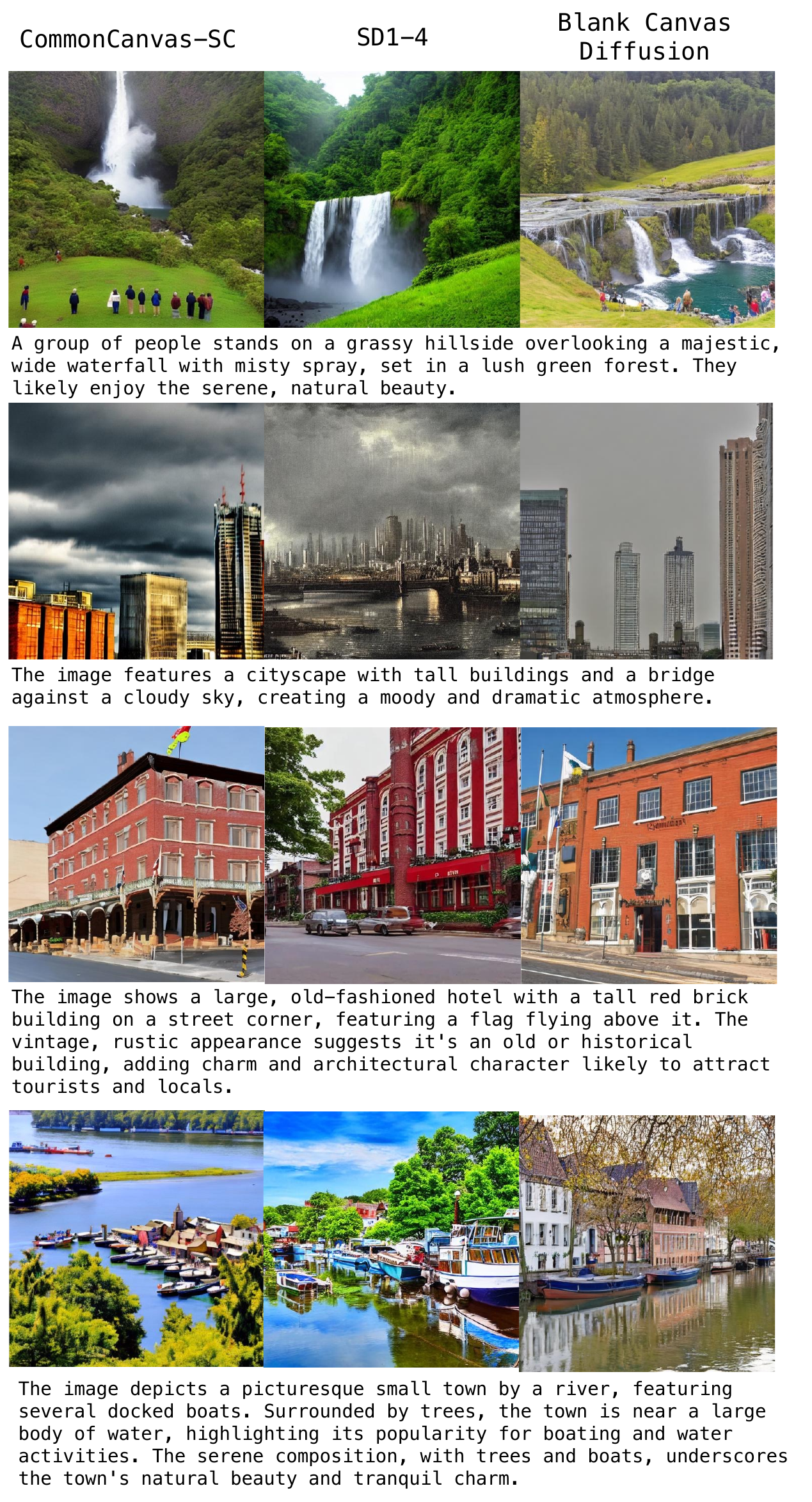}
    \caption{Qualitative comparison of images generated with Blank Canvas Diffusion, Stable Diffusion 1.4 and CommonCanvas-SC model.}
    \label{fig:artfree_gen}
\end{figure}

We demonstrate qualitative results of the Blank Canvas Diffusion in Fig.\ref{fig:artfree_gen}, for comparison, we also include images generated by StableDiffusion 1.4 and CommonCanvas-SC. Our model, despite significantly smaller training set size generates high-quality images faithful to the text prompt.
\vspace{-0.15in}

\section{Qualitative Results of Image Stylization in style of Van Gogh}
\label{sup:qual_imgstyl}

\begin{figure*}[ht]
    \centering
    \includegraphics[width=1\linewidth]{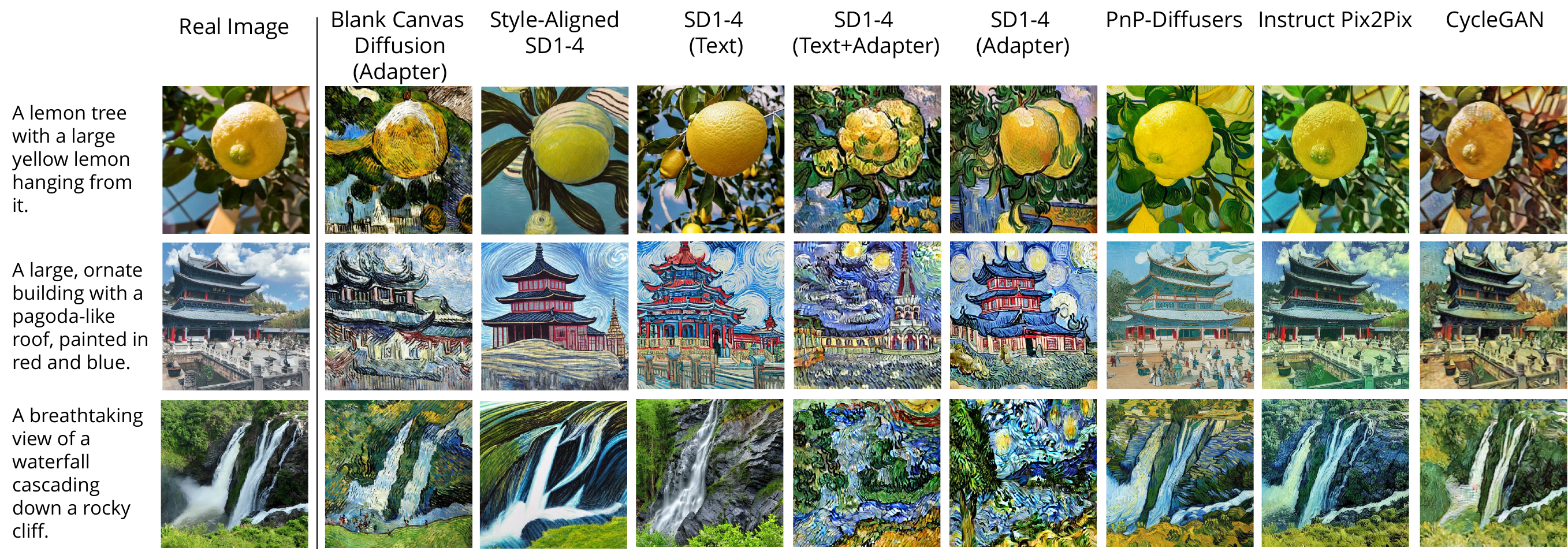}
    \caption{Comparion of our method and other image stylization baselines for the artist Van Gogh. All captions contain a suffix “in the style of Vincent van Gogh", Our model and SD1.4 + Art Adaptor are prompted with suffix “in the style of V* art".}
    \vspace{-0.1in}
    \label{fig:main_style_transfer}
\end{figure*}

\section{Implementation Details}
\label{sup:implementation_details}
As detailed in Sec. 6.1, we train both the VAE and the U-Net from scratch, while utilizing a pre-trained BERT text encoder. 
We train VAE with the mixture of Blank Canvas dataset (SAM) and (filtered) MS-COCO to improve VAE reconstruction quality. While SAM prevents artistic bias, MS-COCO adds non-artistic diverse images. We used 115,294 MS-COCO samples and 104,145 SAM samples. 
We train the VAE with a batch size of 24, gradient accumulation of 2, and a 2e-4 learning rate for 15 epochs, taking ~16 hours.
We first train the U-Net under 256 resolution on 7 H100 GPUs, with each GPU using a batch size of 300 and mixed precision of FP16. We apply gradient accumulation of 8 and use a learning rate of 1e-4 with the AdamW optimizer on a 7 H100 GPUs by 41400 steps. We fine-tune the model at a 512x512 resolution for a total of 156,700 steps, with learning rate of 5e-5 and batch size of 90, and apply 10\% dropping rate with classifier-free guidance sampling~\citep{ho2022classifierfreediffusionguidance}.
The architecture design choices are well motivated to prevent art knowledge leakage from CLIP embedding, otherwise our architecture follows an established baseline SD1.4.
\label{sup:lora_rank}

\begin{figure}[ht]
    \centering
    \includegraphics[width=0.9\linewidth]{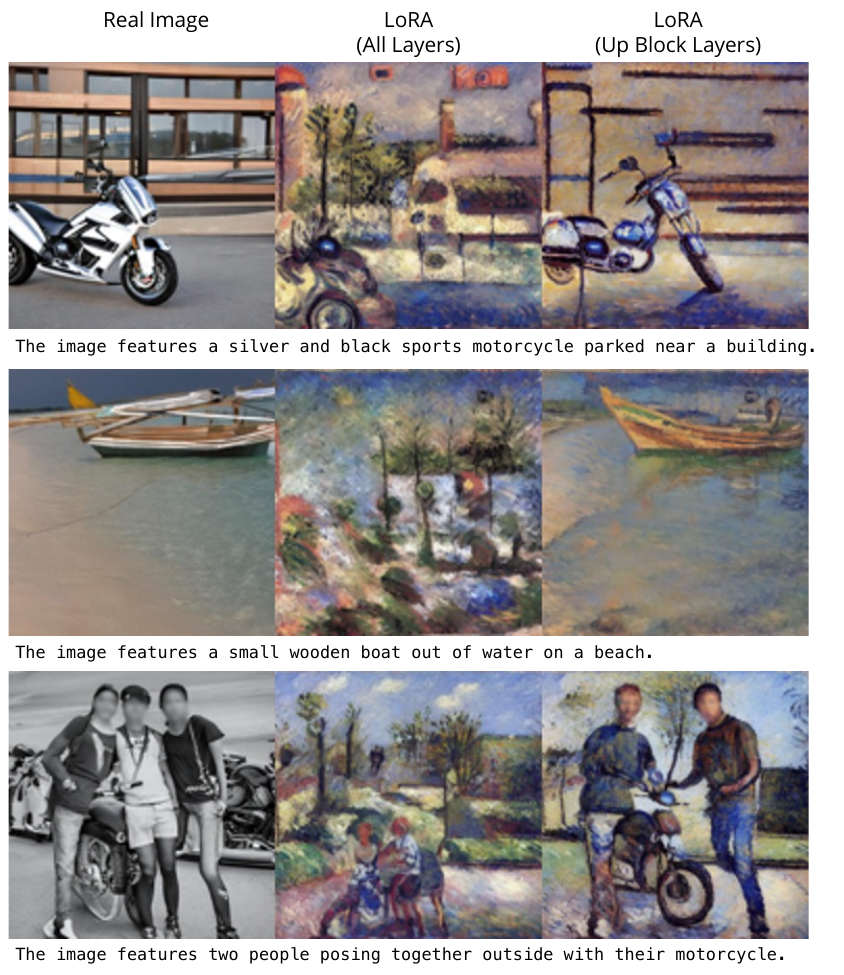}
    \caption{Comparison of LoRA applied to all layers vs. only the up block of the UNet. Limiting LoRA to the up block reduces overfitting. Adapters train on a 10 images sample of Camille Pissaro's artwork.}
    \label{fig:lora_up_full}
    \vspace{-0.1in}
\end{figure}

\paragraphcustom{LoRA Implementation}
We found that incorporating low-rank Adapters into the attention, linear, and convolution layers of the UNet's up block reduces overfitting and improves generation quality, as opposed to introducing LoRA across all UNet blocks.
Motivated by the observation that early layers handle global image aspects, which are less style-dependent, we found that injecting LoRA layers only in the UNet’s up block reduces overfitting (Fig. \ref{fig:lora_up_full}). Quantitative evaluation shows this approach achieves a higher style score across 17 artists (0.29 vs. 0.26) while preserving a comparable content score (0.22 vs. 0.23).
The learning rate was set to 2e-4 using the AdamW optimizer, and we trained for 1,000 steps with a batch size of 5 and the DDIM noise scheduler.
For data augmentation, we resize images with a random scale of 0.9 to 1 and randomly crop with an aspect ratio of 3/4 to 4/3. In the experiments, we use `sks' as the V* token, which serves as a random new token for learning a new art style concept.

Additionally, we conducted an analysis to determine the effect of LoRA rank on the art adapter's performance. Tab.~\ref{tab:ablation:rank} presents the results of our model with LoRA ranks 1 and 64. Our findings indicate that LoRA rank does not significantly impact model performance. This experiment is done on the image stylization task, the scores are average across 17 artists, with 1.0 LoRA scale.

\begin{table}[ht]
    \centering
    \resizebox{0.7\linewidth}{!}{
\begin{tabular}{c|cccc}
    \toprule
    LoRA Rank & CSD$\uparrow$ & LPIPS$\downarrow$ & ViTc$\uparrow$ & CLIPc$\uparrow$ \\ 
    \midrule
    1  & 0.29 & 0.62 & 0.28 & 0.22 \\
    64 & 0.21 & 0.59 & 0.32 & 0.25 \\
    \bottomrule
\end{tabular}
    }
    \caption{Rank analysis of LoRA on style transfer task. We find that a higher rank of LoRA does not improve the model learning performance.}
    \label{tab:ablation:rank}
\end{table}

In our quantitative evaluation, we use a 500-sample subset of the LAION Pop dataset, randomly sampled while excluding images with keywords listed in \ref{sup:filtering}. For captions longer than a baseline’s content length, we use only the first sentence.

\paragraphcustom{Content Loss Strength}
\label{sup:prior_ablation}
We investigated the influence of the content loss weight ($w$) in the art adapter across different models Tab. \ref{tab:ablation:preservation}.

The content loss substantially enhances learning performance, with CSD increasing from 0.14 to 0.29 when $w$ is set to 50. This demonstrates that the content loss effectively aids the model in distinguishing between art images and natural images. The effect remains robust across different weight values, with performance remaining nearly constant when $w$ is set to 20 or 100 (up to 0.02 difference in CSD).

\begin{table}[ht]
\centering
    \vspace{-0.5em}
    
\resizebox{0.8\linewidth}{!}{
    \centering

\begin{tabular}{c|c|c|c|c}
    \toprule
    Content Loss scale & CSD$\uparrow$ & LPIPS$\downarrow$ & ViTc$\uparrow$ & CLIPc$\uparrow$ \\ 
    \midrule
    0    & 0.14	&0.57	&0.33	&0.25 \\ 
    20   & 0.29	&0.62	&0.28	&0.22 \\ 
    50   & 0.29	&0.62	&0.28	&0.22 \\ 
    100  & 0.27	&0.61	&0.28	&0.23 \\ 
    \bottomrule
\end{tabular}
}
\caption{Analysis of prior preservation loss weight ($w$) on our model. Experiments are conducted on style transfer, with noise added at the 800th time step. The scores are averages across 17 artists on image stylization task, with 1.0 LoRA scale.}
\label{tab:ablation:preservation}
\end{table}

\section{Art-Agnostic Model Verification}
\label{sup:art_agnostic}

\begin{figure}[htbp]
    \centering
    \includegraphics[width=0.7\linewidth]{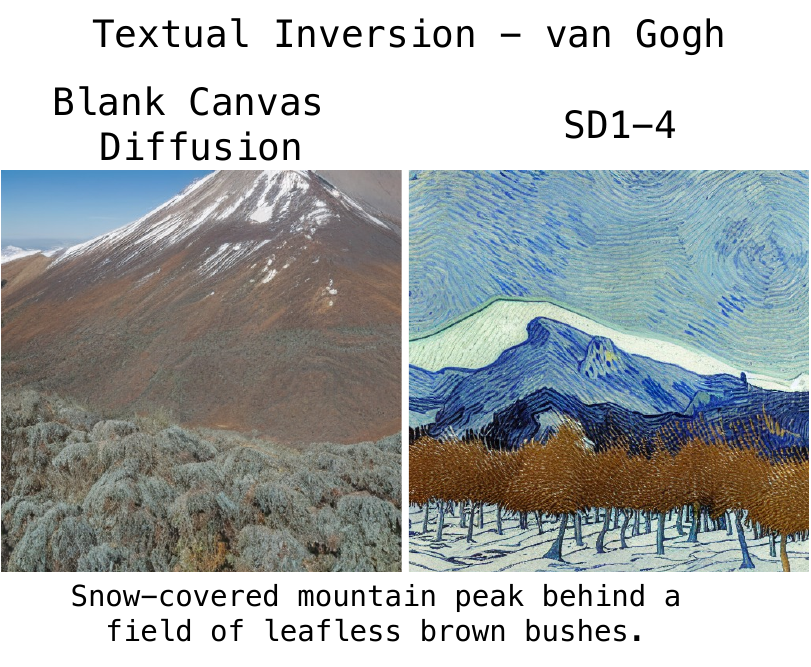}
    \caption{Through textual inversion using paintings by van Gogh, we found that, unlike SD1-4, our model cannot generate images in the corresponding style. This indicates that our model cannot be inverted to generate artwork through prompt space searching, demonstrating it has no prior knowledge of art.}
    \label{fig:textual_inversion}
\end{figure}

To verify the art-agnostic nature of our model, we conducted a textual inversion experiment as suggested by \citet{pham2023circumventing}. In the experiment we use the same Art Dataset as for training the Art Adapter for Vincent Van Gogh style. Fig. \ref{fig:textual_inversion} illustrates that our model fails to produce the target style using textual inversion, further confirming its lack of prior artistic knowledge.

\section{Model Editing and Controlling Ability}
\label{sup:editing}
Despite being trained on a significantly smaller and less diverse dataset limited to natural images, our art-agnostic model demonstrates comparable editing and control capabilities to competitive models. This is evident in both single-image editing and customization experiments.

\begin{figure}[htbp]
    \centering
    \includegraphics[width=\linewidth]{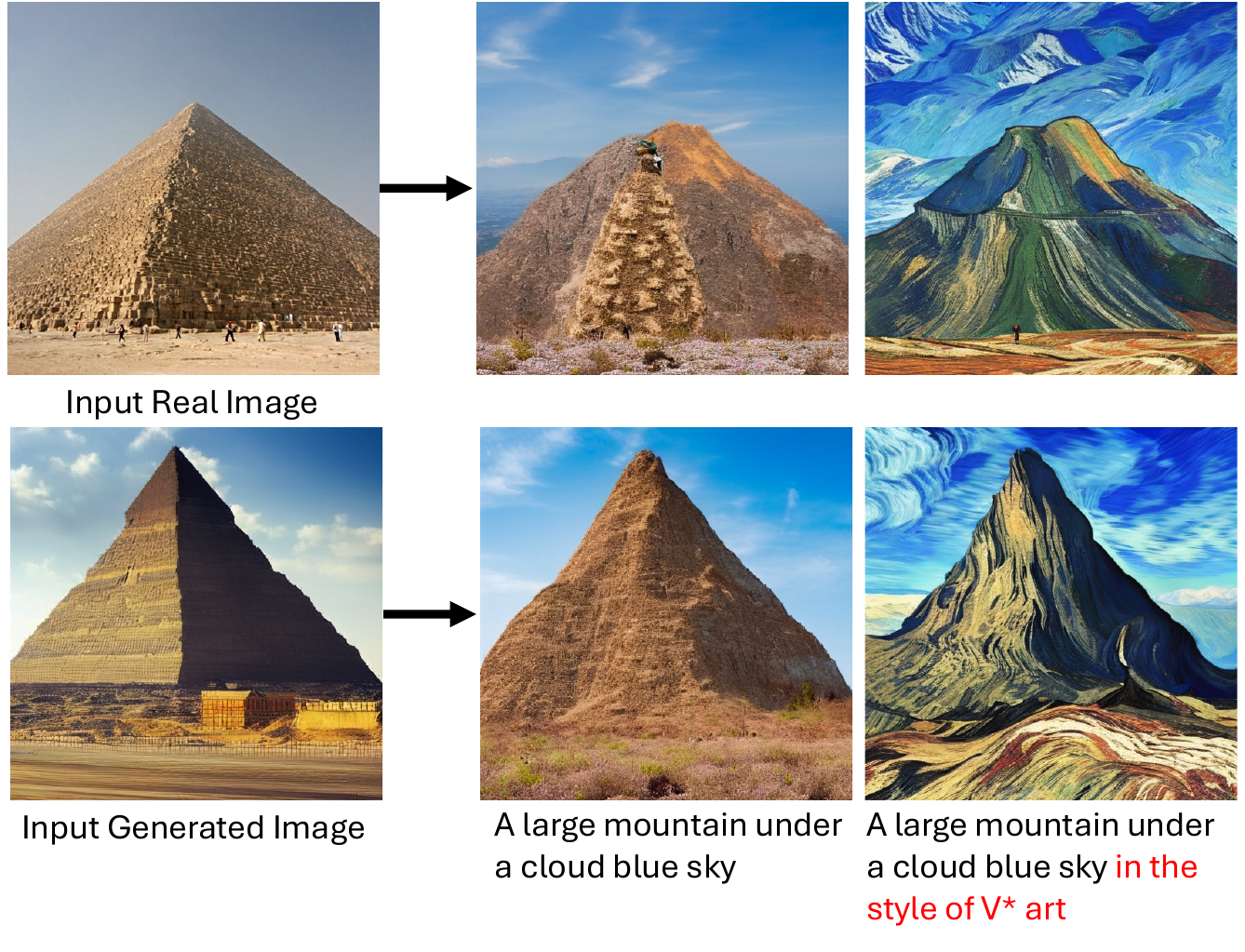}
    \caption{Plug-and-Play editing on our model. We provide both editing on real and generated image examples. We replace a pyramid to a large mountain both without and with the artistic adaptor of van Gogh.}
    \label{fig:pnp}
\end{figure}

In Fig. \ref{fig:pnp}, we qualitatively illustrate the single-image editing process using the Plug-and-Play method \citep{Tumanyan_2023_CVPR} applied to our model. We provide editing examples on both real and generated images, demonstrating the model's ability to replace a pyramid with a large mountain, both with and without the artistic adapter (weight 1.5) of van Gogh.

Furthermore, we demonstrate our model's customization abilities using the Dreambooth technique \citep{ruiz2023dreamboothfinetuningtexttoimage}. We learned the concept of a barn using 7 training images from the CustomConcept101 dataset \citep{kumari2023multiconceptcustomizationtexttoimagediffusion}. The model was trained to generate the barn in various contexts, utilizing 200 prior samples from Stable Diffusion v1-4, with a prior preservation loss of 1.0, a learning rate of 5e-6, and 250 training steps on 2 GPUs.

\begin{figure}[htbp]
    \centering
    \includegraphics[width=\linewidth]{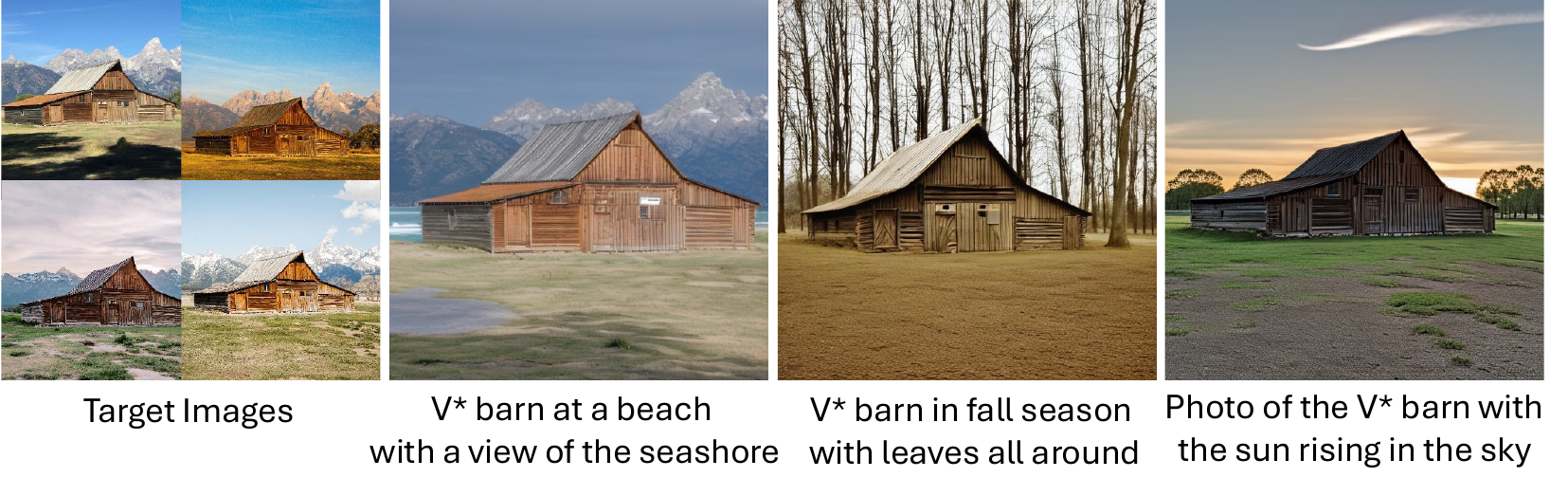}
    \caption{Dreambooth editing on our model. We send 7 barn example images to the model and ask it to generate the barn in various contexts.}
    \label{fig:dreambooth}
\end{figure}

\section{Effect of Applying the Adapter at various Time Steps}
\label{sup:adapter_effect}

\label{sec:supmat:adaptor_time_step}
\begin{figure}[htbp]
    \centering
    \includegraphics[width=\linewidth]{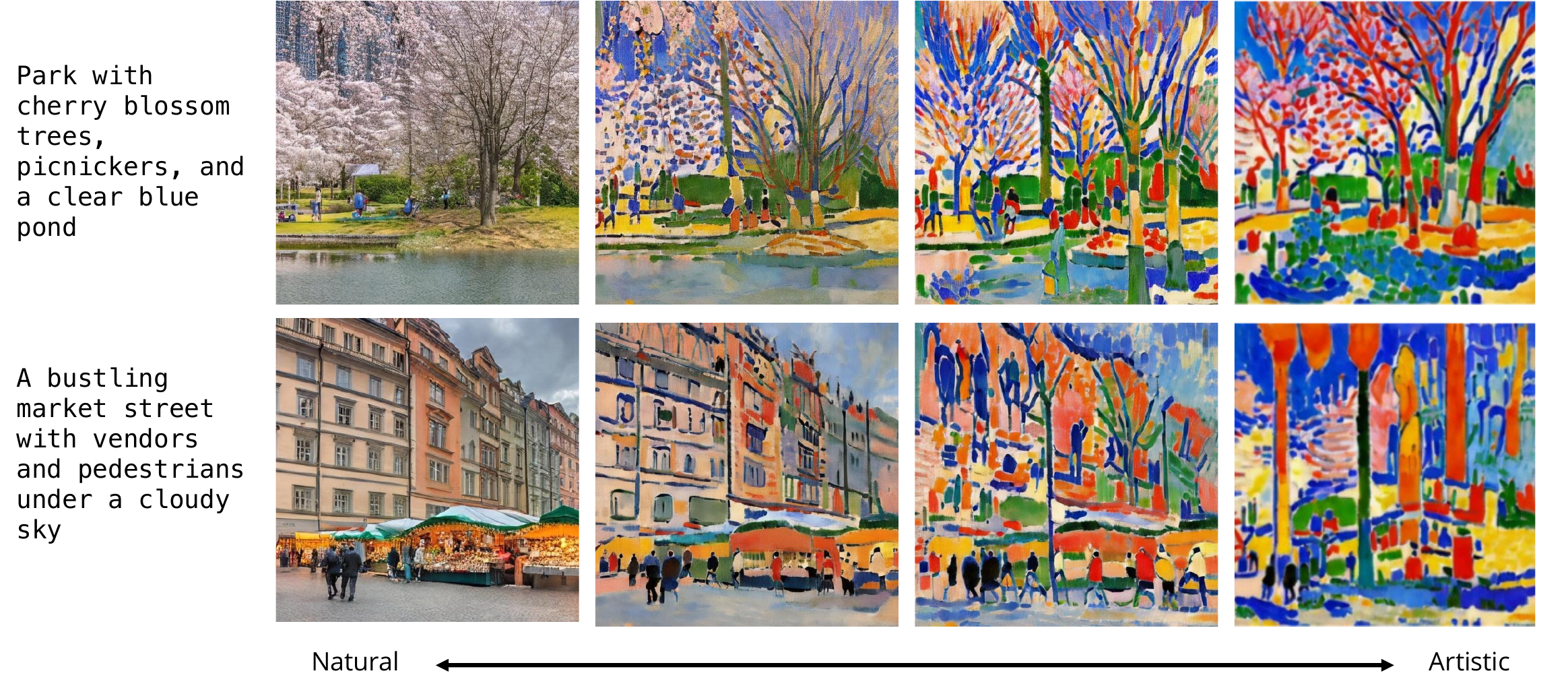}
    \caption{Art generation results using Art Adapter at different timesteps. From left to right: no adapter (column 1), adapter introduced at timestep 800 (column 2), 600 (column 3), and 0 (column 4). This demonstrates how earlier adapter introduction increases artistic influence in the image.}
    \label{fig:adaptor_step_derain}
\end{figure}

We analyzed the effect of the adapter time step on art generation results. Fig. \ref{fig:adaptor_step_derain} shows the art generation outcomes with different adapter time steps. Intuitively, the model generates more style information when the adapter starts earlier (left) and more content information when the adapter starts later (right).

\section{User Study}
\label{sup:user_study}

\begin{figure}[htbp]
    \centering
    \includegraphics[width=\linewidth]{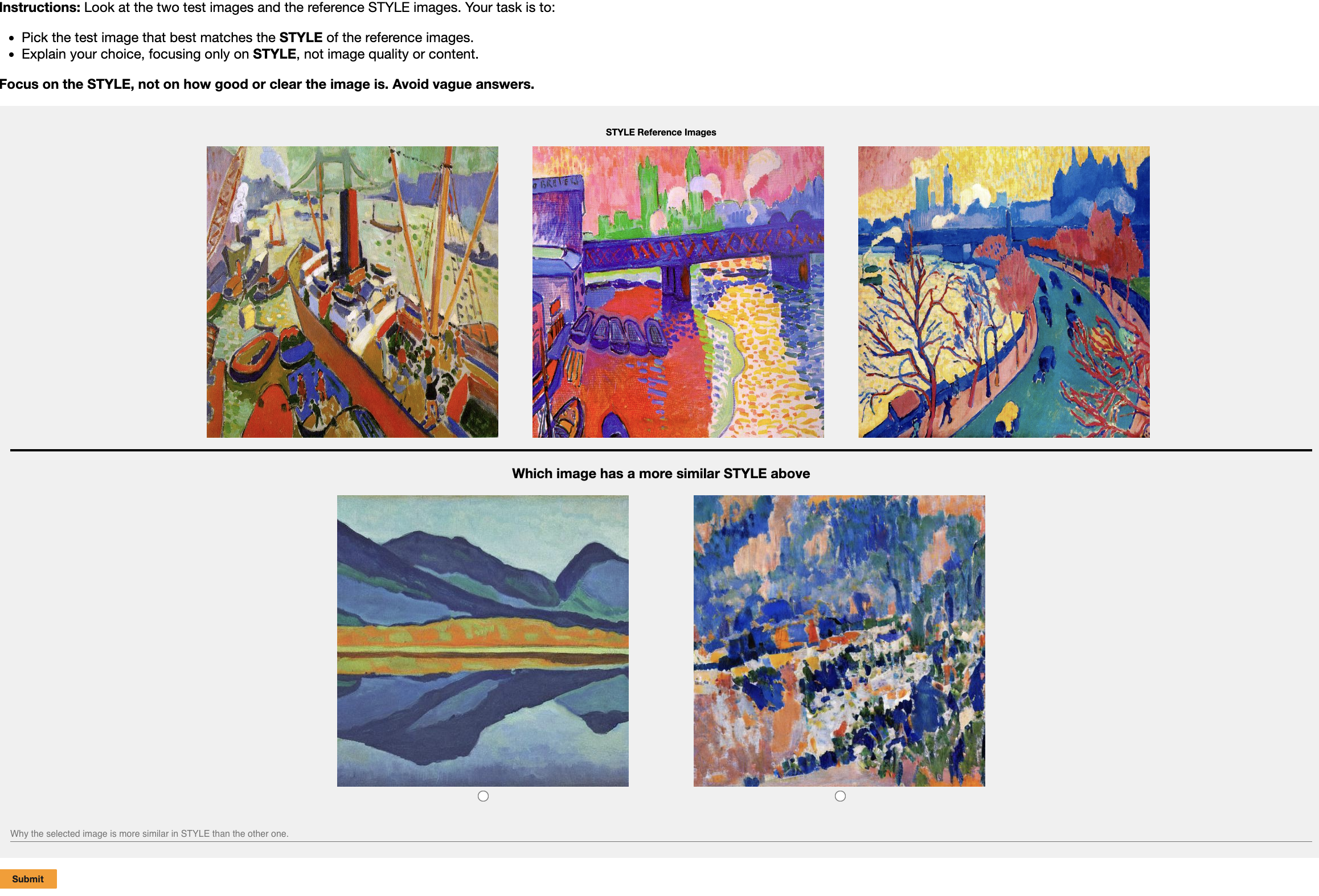}
    \caption{The interface for the user study we conducted, the participants were shown examples of an art style and were asked to select an image that matched it better.}
    \label{fig:user_study_interface}
\end{figure}

The user study was conducted on AWS with 42 participants. We disregarded the workers that took less than 5 seconds to complete the assignment. To ensure reliability, we included a validation task requiring participants to distinguish a non-artistic image from a painting, ensuring they focused on stylistic qualities rather than content or quality of the images. Only those who passed this task were included in the analysis. This process reduced the number of workers from an initial pool of 88 to 42.
Additionally, we selected expert workers and randomized both the image order and methods to minimize potential bias. The user study interface is shown in Fig. \ref{fig:user_study_interface}.

\section{Introducing the Art Adapter to the Artist}
\label{sup:alan}

To explore the artistic community's reaction to AI-generated art, we conduct an interview with the renowned artist Alan Kenny. Upon obtaining Alan's permission, we train an Art Adapter on 11 artworks showing his distinctive style. We describe our work and present Kenny with the generated images imitating his style.
In the interview, the artist expresses a blend of astonishment and familiarity when observing the AI-generated art, remarking, ``I didn’t expect [this quality] if you were using a base model of blank canvas... you probably achieved more than I would have expected for a base model with no information.'' He acknowledges that the AI has captured aspects of his distinct style to the extent that, ``if you were to post some of these images online, I would get people texting me, ‘I see your images.’ They would spot it, and I spot it.'' Despite noting that ``compositionally, it is weak'' and contrasting this with his own ``well thought and meticulous'' compositions, he recognizes that ``there are some very positive things'' in the AI’s work.
The artist describes the experience as ``terrifying and a bit exciting at the same time,'' specifically pointing out how the AI imitates his signature ``gradation of the landscape'' and ``gradation of the shapes.'' Though he felt his style is largely captured, he admits, ``there is kind of originality to them... I see me in them, yes, very strongly... but there is an originality to some of the images.''.

\begin{figure*}[htbp]
    \centering
    \includegraphics[width=\linewidth]{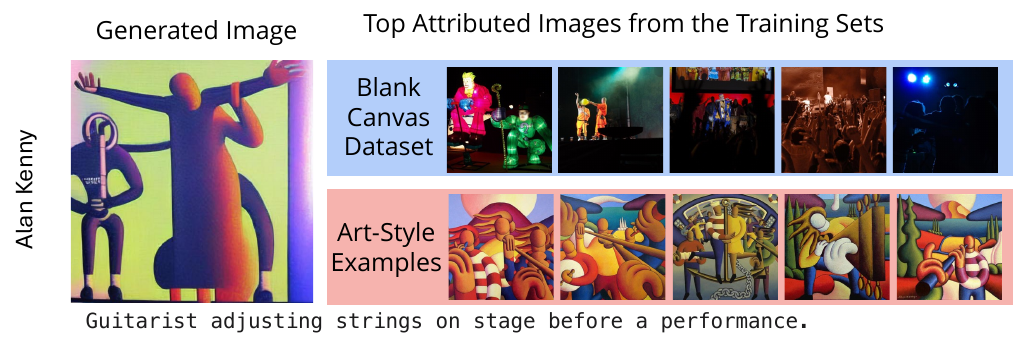}
    \caption{Generated artwork in the style of Alan Kenny (created and displayed with the artist's permission) showcases the top-5 influential images from the Blank Canvas and Art Datasets.}
    \label{fig:alan}
\end{figure*}

\begin{figure*}[htbp]
    \centering
    \includegraphics[width=\linewidth]{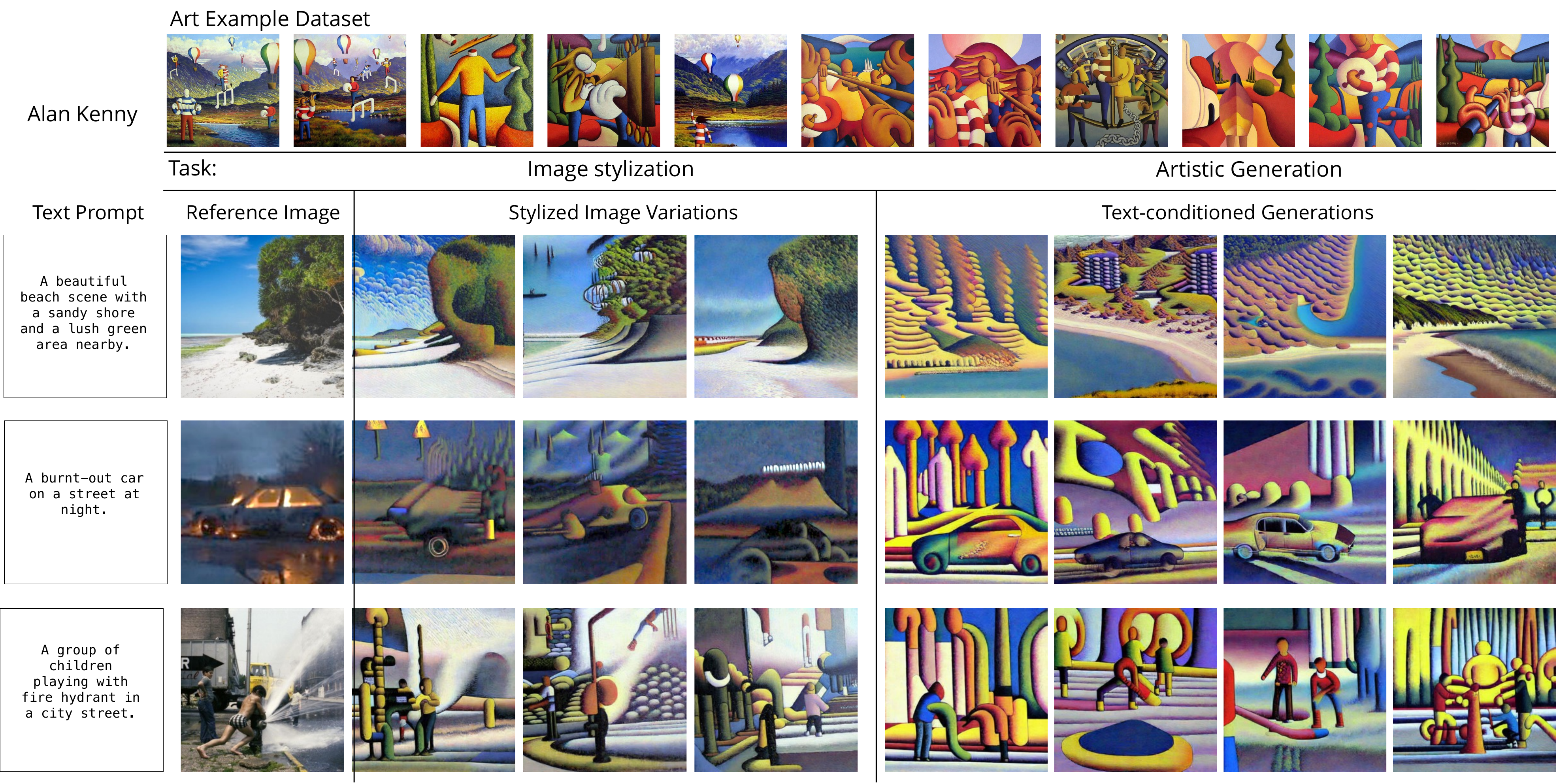}
    \caption{Additional qualitative experiments of the art imitation of the interviewed artist Alan Kenny.}
    \label{fig:q:alan}
\end{figure*}

In Fig. \ref{fig:alan}, \ref{fig:q:alan} we present qualitative examples of images generated in the style of Alan Kenny along with the results of the data attribution technique. These examples reveal how natural images inspire features in the generated art (e.g., a stage with musicians) while preserving the characteristics of the artistic style like the use of colors, smooth boundaries and geometric shapes.

\section{Data Attribution}
\label{sup:attribution}

\begin{figure*}[htbp]
    \centering
    \includegraphics[width=0.75\linewidth]{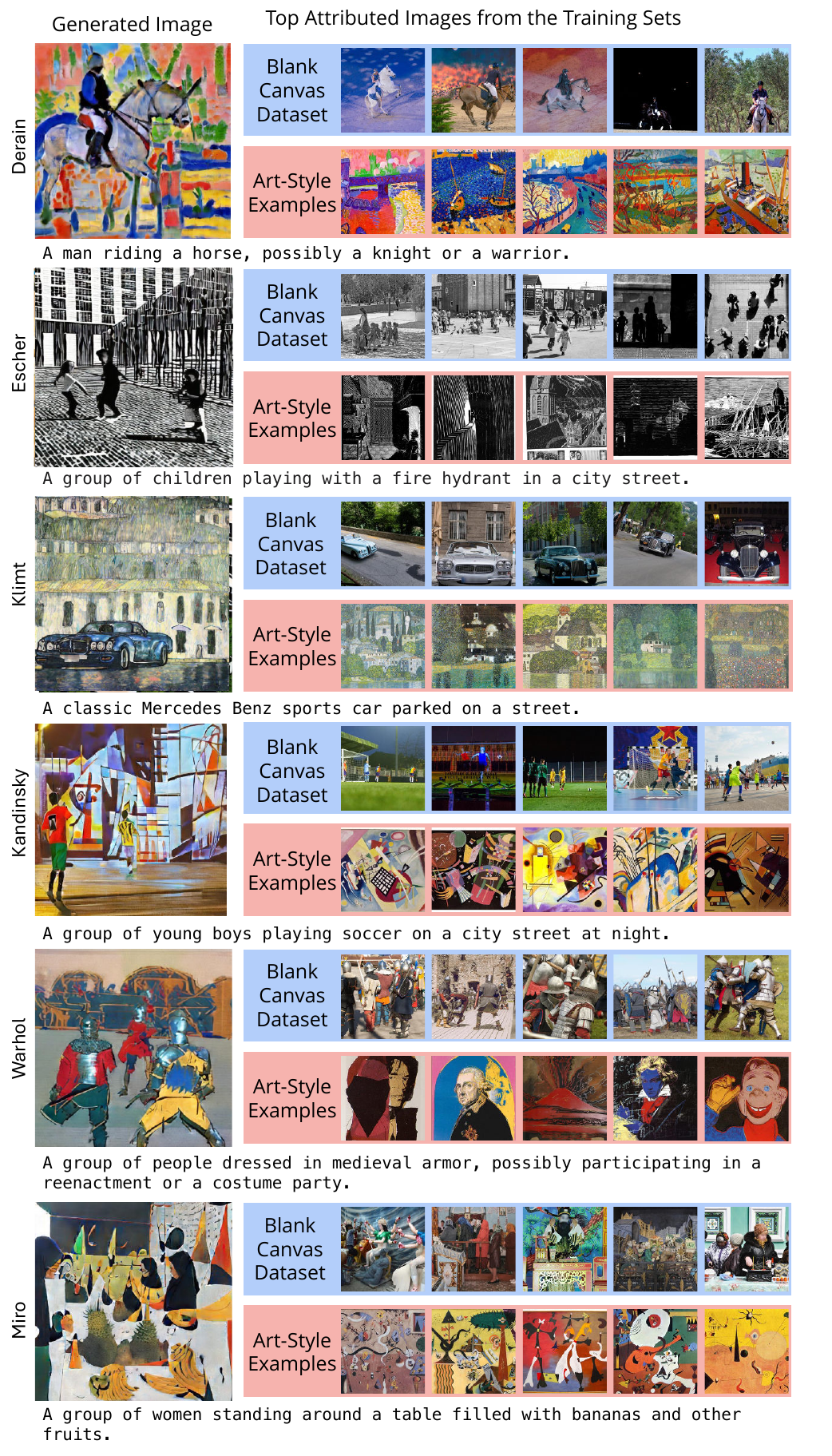}
    \caption{Additional qualitative experiments showing diverse art generations and top five attributed images from both the Blank Canvas dataset and and Art-Style example dataset.}
    \label{fig:attr_sup}
\end{figure*}

We present additional results of applying data attribution to the generated art images in Fig. \ref{fig:attr_sup}. These results illustrate how specific visual elements from the training data, including both natural and small art images, influence the generated outputs. Despite the Art Adapters being trained on a limited set of art images, and the base text-to-image model itself having minimal exposure to graphic art, the attribution analysis points to similarities in the natural images that may enable the model to effectively generalize from few examples. 

The attribution method [47] is an established approach, with the similarity metric chosen to closely approximate exact attribution. Although the method is an approximation, no current data attribution method can pin-point training examples more precisely at the scale of training data we need to analyze. We acknowledge this gap and will add the discussion to the main paper.

\section{Different Baselines}
\label{sup:baselines}

While many methods transfer style from a reference image to another, direct comparisons are often infeasible due to differences in model architecture and dependencies. For instance, StyleDrop \cite{sohn2023styledrop} is designed specifically for the Muse architecture, making it difficult to separate the contribution of the adaptation method from the pretrained model’s inherent stylization capabilities. Similarly, Visual Style Prompting \cite{jeong2024visual} and InstantStyle \cite{wang2024instantstyle} are designed primarily for Stable Diffusion XL.
Computational constraints prevent us from training a comparable model with our Blank Canvas data, but we encourage others to explore similar experiments.

DeadDiff \cite{qi2024deadiff}, while offering advantages to text-to-image adapters, relies on a paired dataset where the reference image and ground truth share style or semantics, which differs significantly from our approach. Our primary goal is to demonstrate that effective style transfer is achievable with a few examples, rather than competing with methods that leverage extensive pretrained knowledge of graphic art.

To disentangle the adaptation method from the pretrained model's capabilities, we applied another baseline, StyleID \cite{Chung_2024_CVPR}, to both our Blank Canvas Diffusion model and SD1.4. Similar to StyleAligned, both training-free adaptation methods performed better on SD1.4, leveraging its broad artistic knowledge, but struggled on Blank Canvas Diffusion, highlighting their reliance on pretrained models rich in artistic priors. In contrast, our Art Adapter bridges this gap effectively, demonstrating that focused adaptations within the Blank Canvas framework can achieve compelling results without relying on inherited artistic biases.

While this comparison is not entirely equivalent—StyleAligned and StyleID use a single reference image, whereas our Art Adapter employs multiple style references (in this experiment, we compare five artists: Derain, Miró, Klimt, Picasso, and Lichtenstein, with an average training set of 15)—we were unable to adapt these methods to support multiple references, as doing so falls outside the scope of this work.

It is important to emphasize that our goal is not to compete with models and methods trained on significantly larger graphic art datasets, as such comparisons would be inherently unfair. Instead, our work focuses on a key question: how much graphic art data is truly needed to effectively replicate an artistic style? Our analysis demonstrates that an artistic style can be successfully learned from just a few examples.

\begin{table}[ht]
    \centering
    \resizebox{0.8\linewidth}{!}{
    \begin{tabular}{c|c|ccc}
        \toprule
        Text-To-Image Model & Adaptation Method & CSD\_mean$\uparrow$ & ViTc$\uparrow$ & CLIPc$\uparrow$ \\ 
        \midrule
        & Art-Adapter & \textbf{0.35} & 0.27 & \textbf{0.23} \\ 
        Blank Canvas Diffusion & StyleAligned & 0.12 & 0.31 & 0.22 \\ 
        & StyleID & 0.11 & 0.63 & 0.22 \\ \cmidrule(lr){1-5}
        & Art-Adapter & 0.21 & 0.27 & \textbf{0.26} \\ 
        SD1.4 & StyleAligned & \textbf{0.43} & 0.23 & 0.21 \\ 
        & StyleID & 0.29 & 0.40 & 0.23 \\ 
        \bottomrule
    \end{tabular}
    }
    \caption{Comparing different art adaptation methods across our Blank Canvas Diffusion model and Stable Diffusion 1.4. Training-free adaptation methods, StyleAligned and StyleID, perform better on SD1.4, benefiting from the model's broad artistic knowledge, but struggle on Blank Canvas Diffusion, showing their reliance on pretrained models rich in artistic priors. In contrast, our Art Adapter effectively bridges this gap, proving that focused adaptations within the Blank Canvas framework can deliver compelling results without depending on inherited artistic biases.}
    \label{tab:model_comparison}
\end{table}

\section{Additional Qualitative Results}
\label{sec:supmat:art_generation}
\begin{figure*}[htbp]
    \centering
    \includegraphics[width=\linewidth]{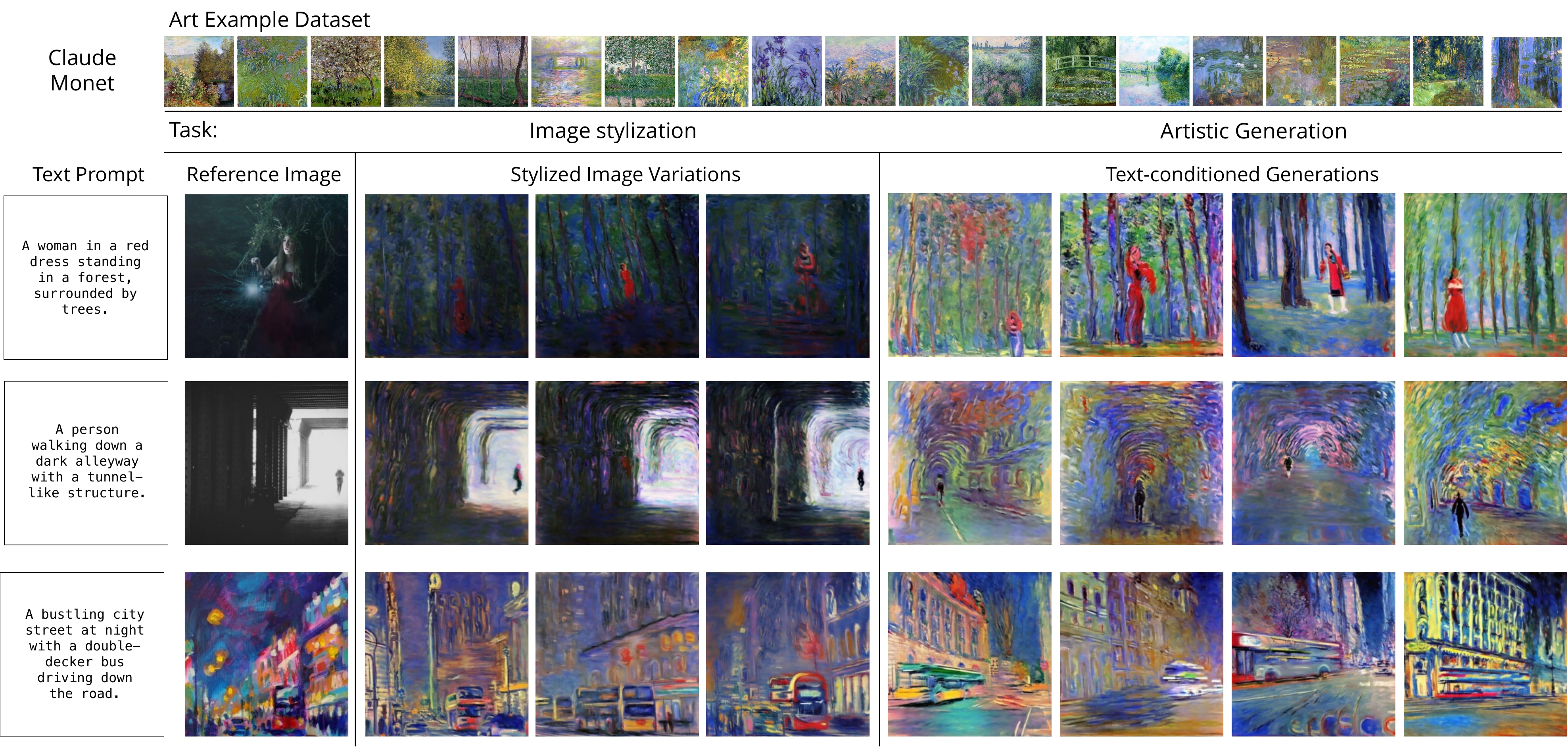}
    \caption{Additional qualitative experiments.}
    \label{fig:q:monet}
\end{figure*}

\begin{figure*}[htbp]
    \centering
    \includegraphics[width=\linewidth]{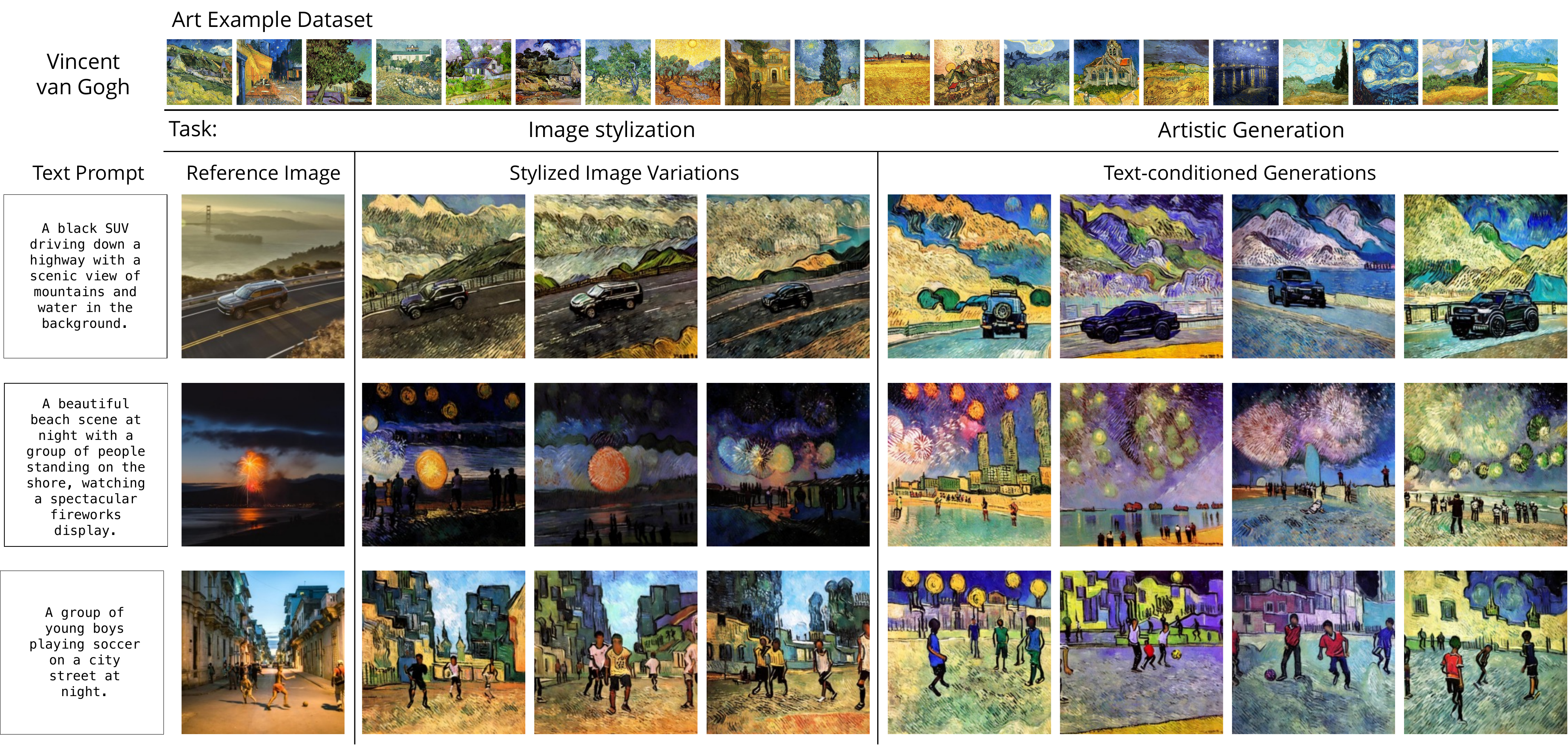}
    \caption{Additional qualitative experiments.}
    \label{fig:q:vangogh}
\end{figure*}

\begin{figure*}[htbp]
    \centering
    \includegraphics[width=\linewidth]{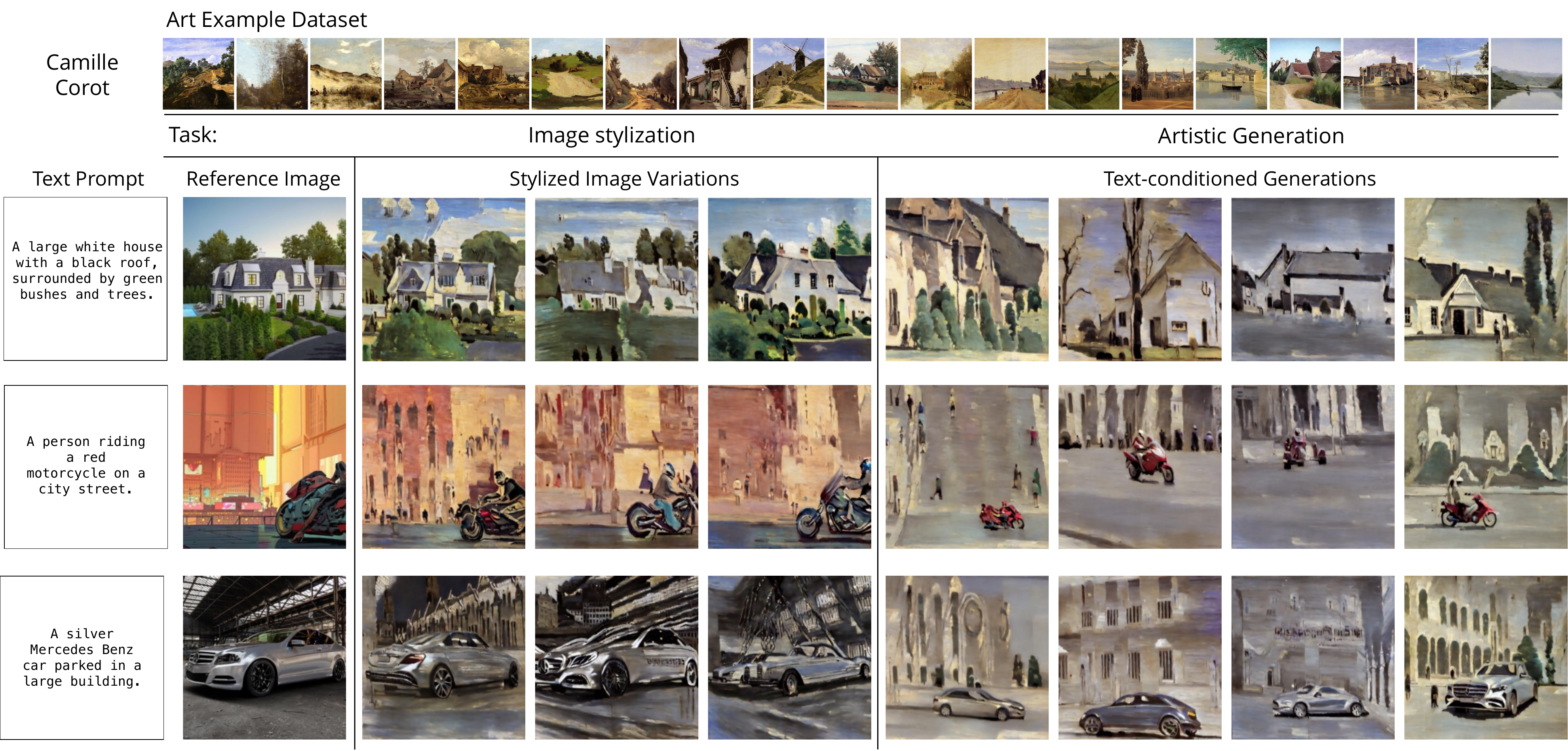}
    \caption{Additional qualitative experiments.}
    \label{fig:q:corot}
\end{figure*}

\begin{figure*}[htbp]
    \centering
    \includegraphics[width=\linewidth]{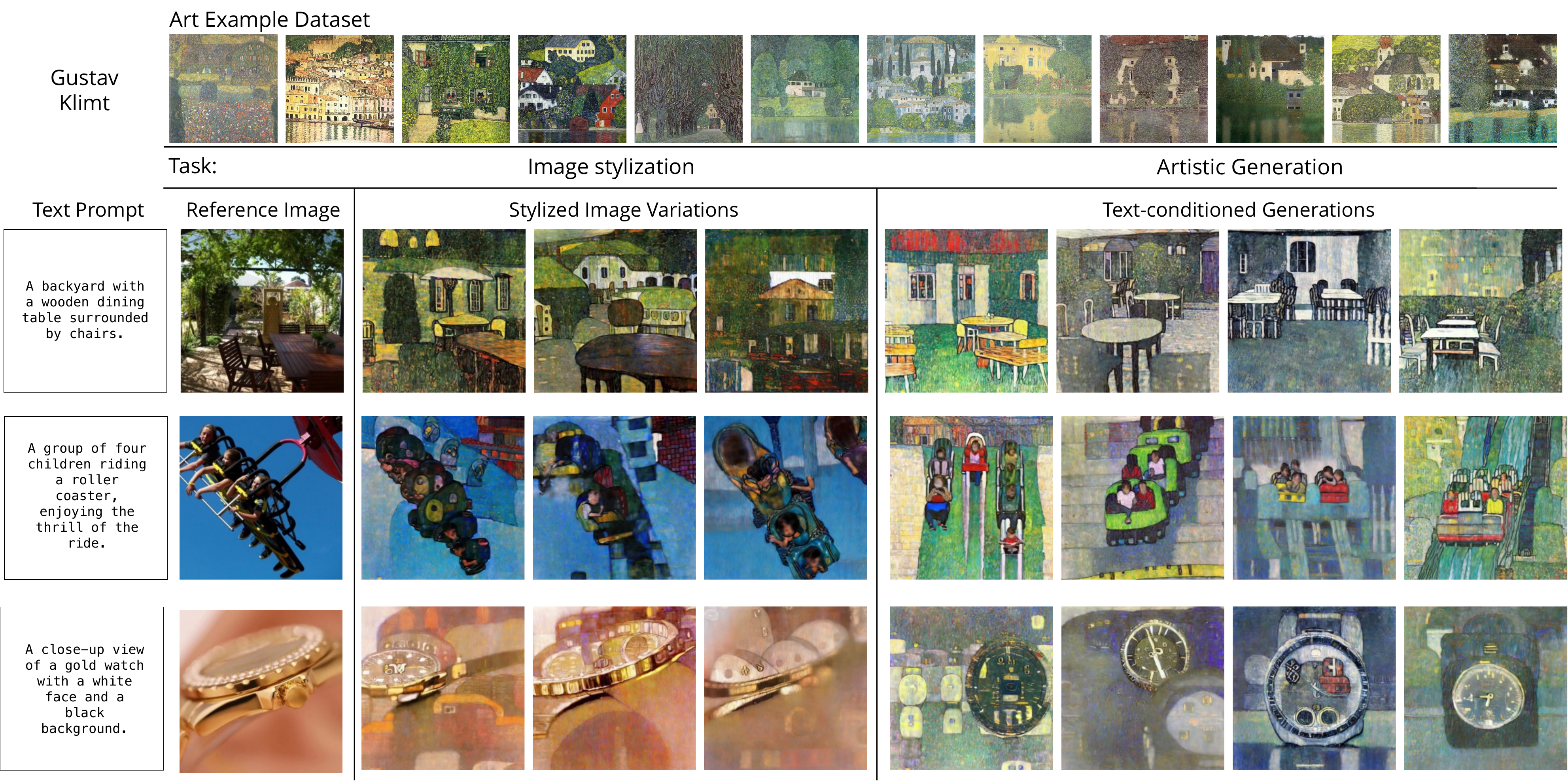}
    \caption{Additional qualitative experiments.}
    \label{fig:q:klimt}
\end{figure*}

\begin{figure*}[htbp]
    \centering
    \includegraphics[width=\linewidth]{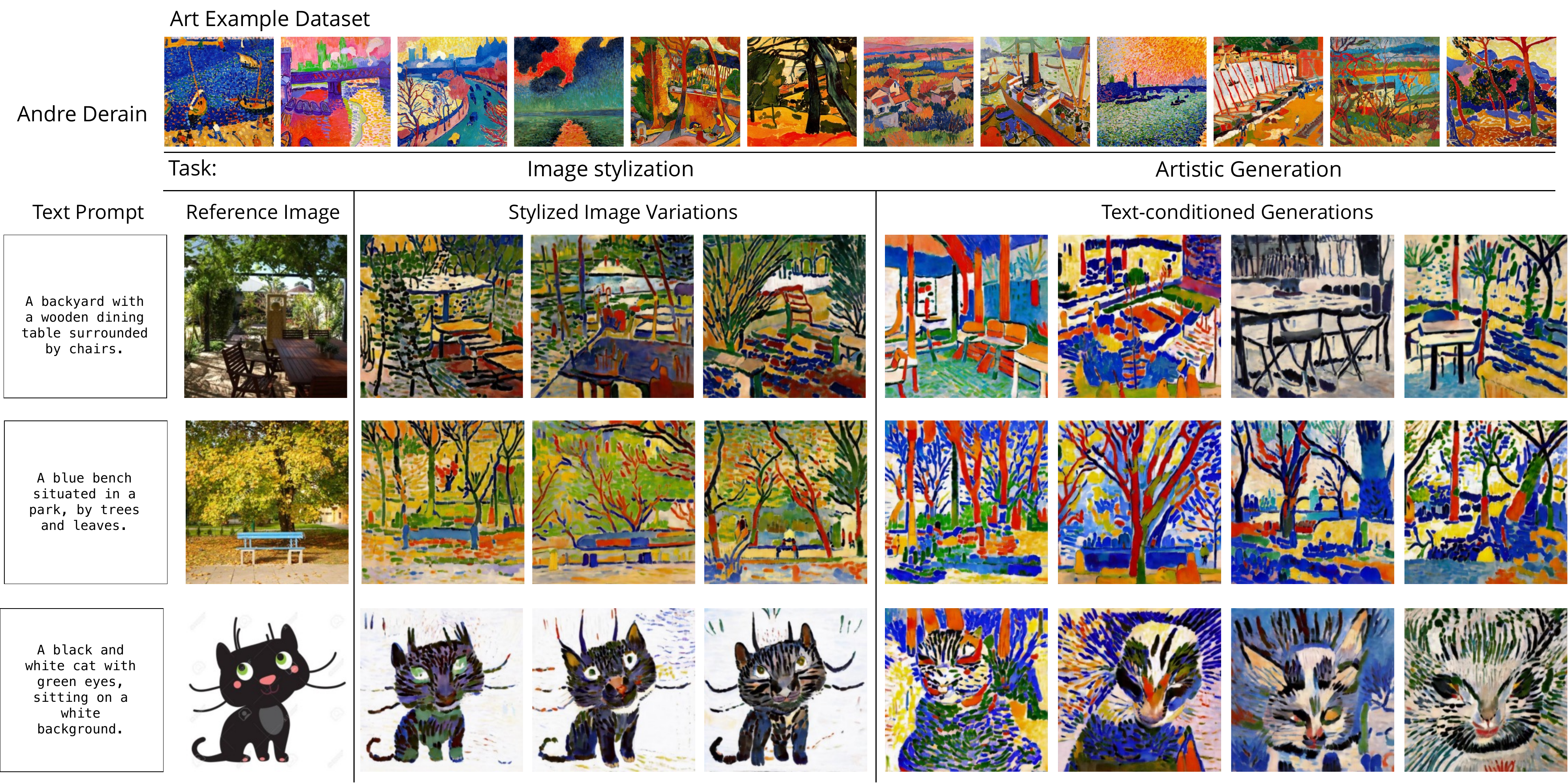}
    \caption{Additional qualitative experiments.}
    \label{fig:q:derain}
\end{figure*}

\begin{figure*}[htbp]
    \centering
    \includegraphics[width=\linewidth]{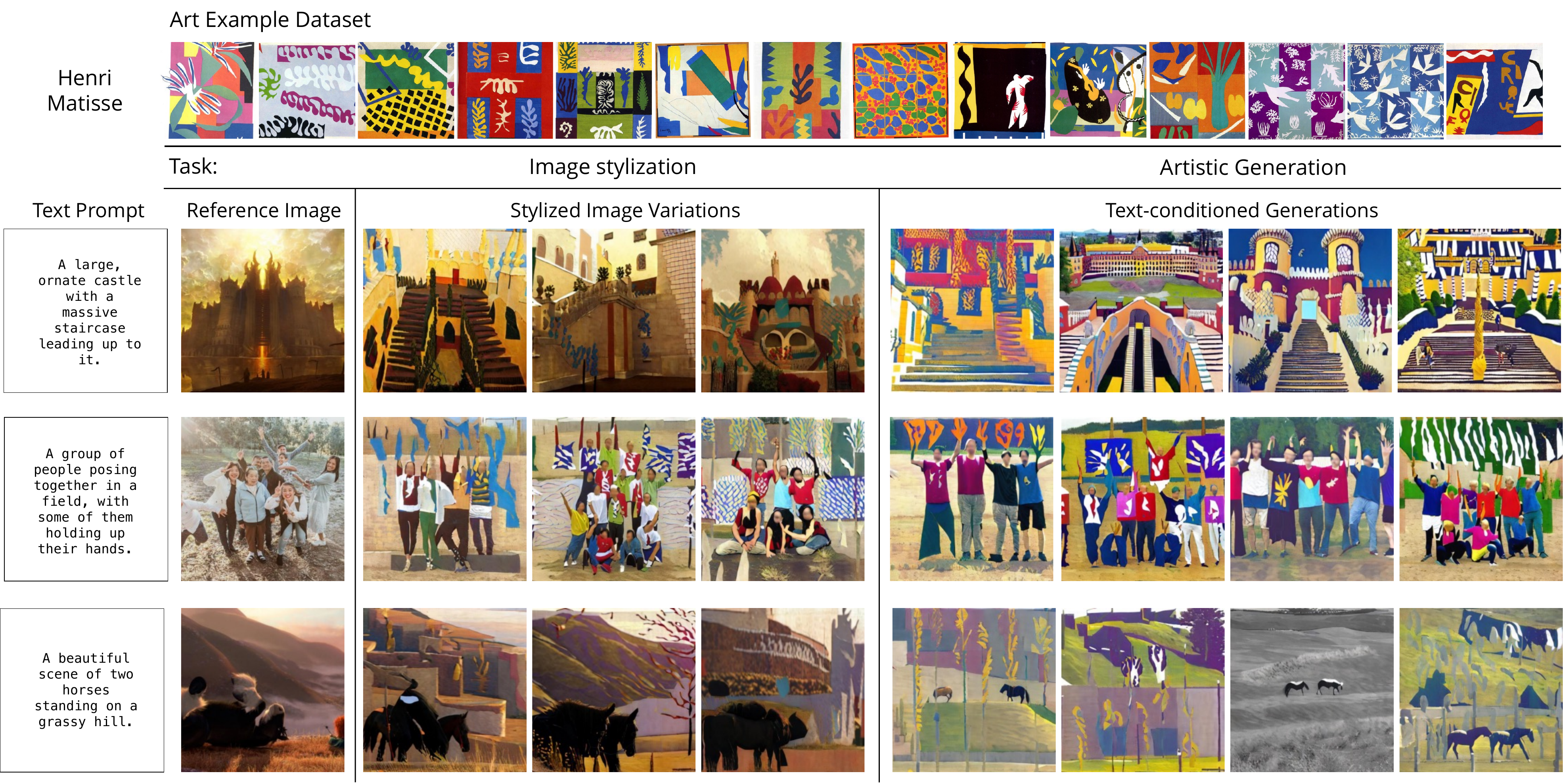}
    \caption{Additional qualitative experiments.}
    \label{fig:q:matisse}
\end{figure*}

\begin{figure*}[htbp]
    \centering
    \includegraphics[width=\linewidth]{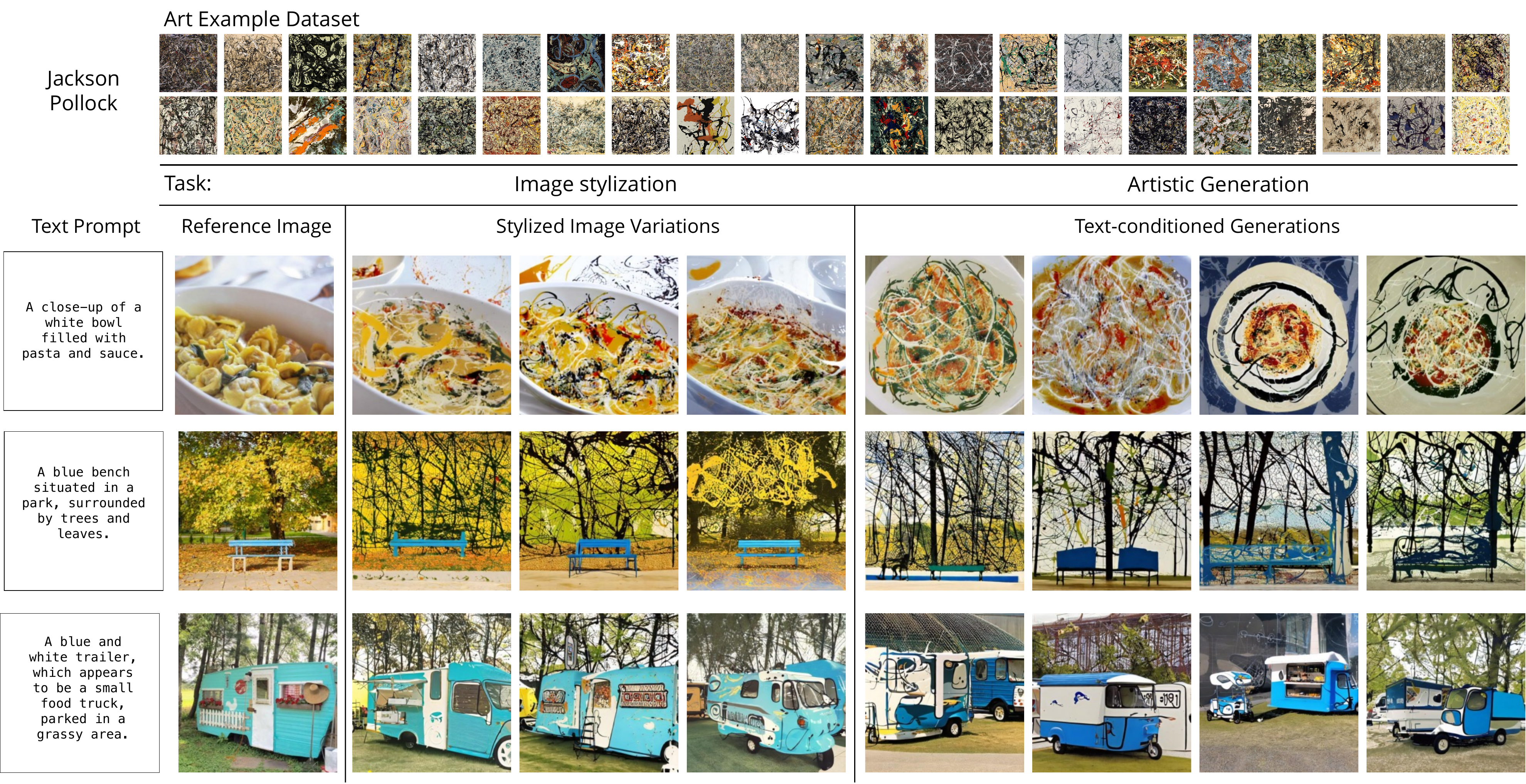}
    \caption{Additional qualitative experiments.}
    \label{fig:q:pollock}
\end{figure*}

\begin{figure*}[htbp]
    \centering
    \includegraphics[width=\linewidth]{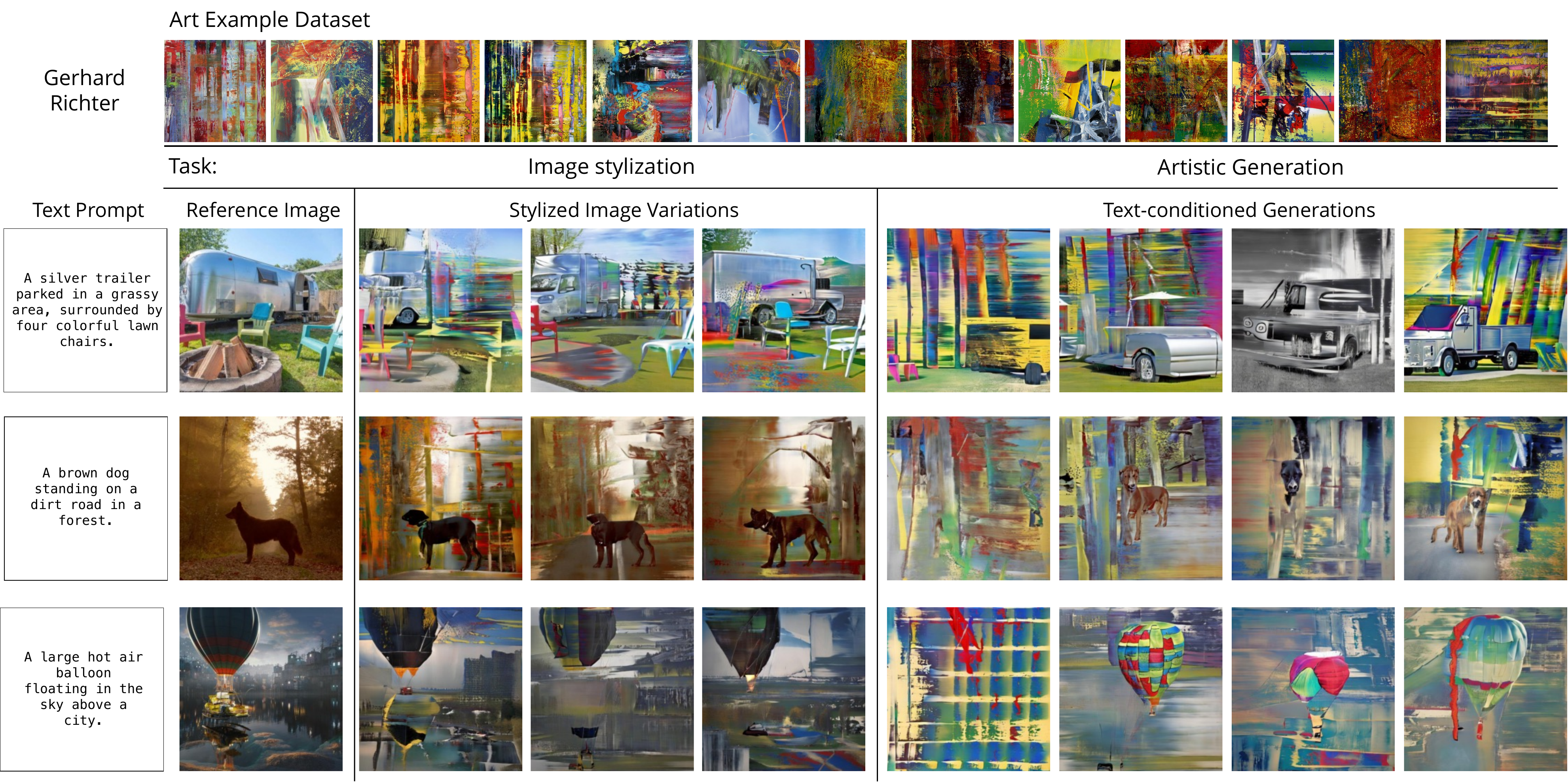}
    \caption{Additional qualitative experiments.}
    \label{fig:q:richter}
\end{figure*}

\begin{figure*}[htbp]
    \centering
    \includegraphics[width=\linewidth]{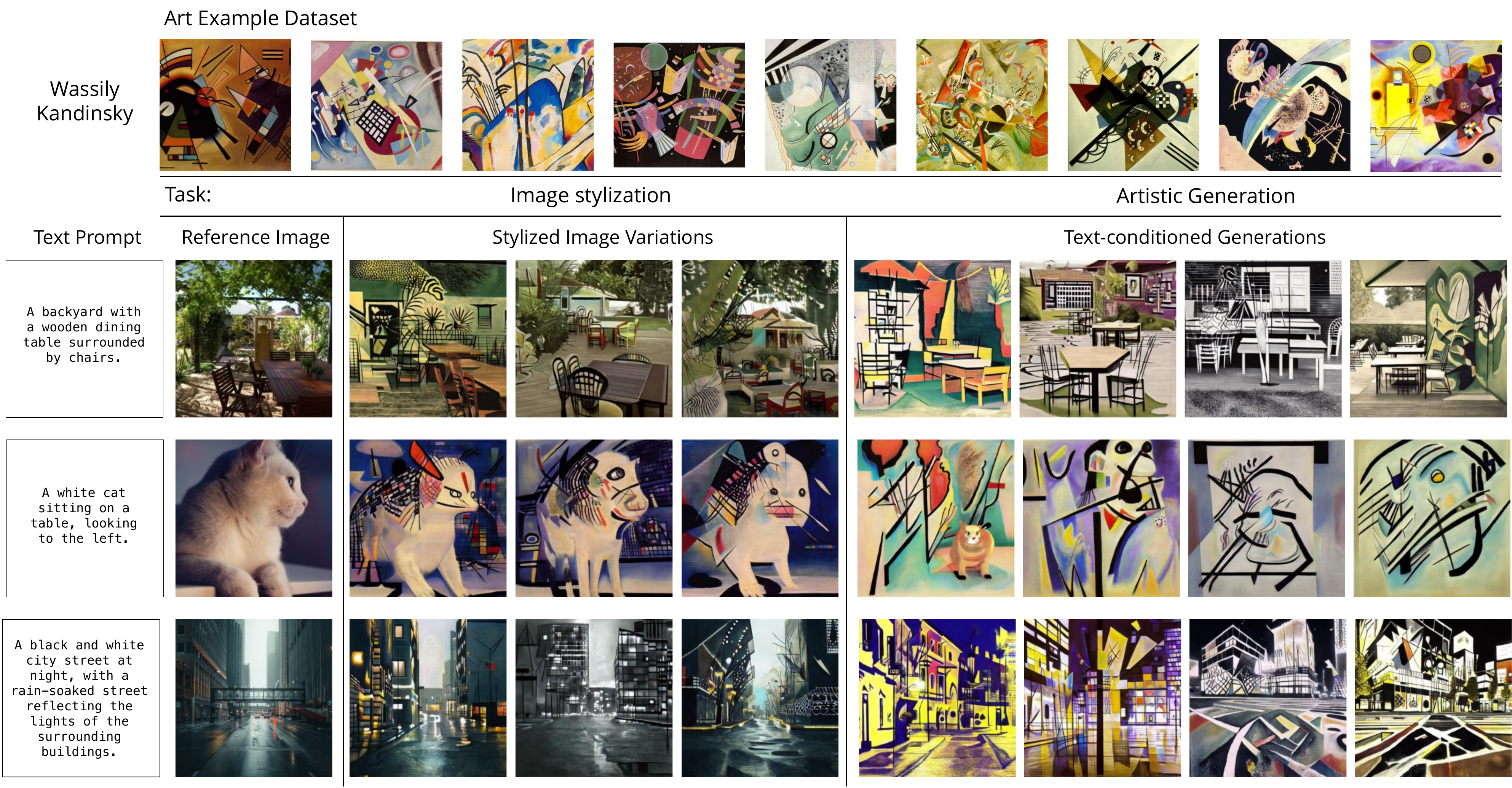}
    \caption{Additional qualitative experiments.}
    \label{fig:q:kandinsky}
\end{figure*}

\begin{figure*}[htbp]
    \centering
    \includegraphics[width=\linewidth]{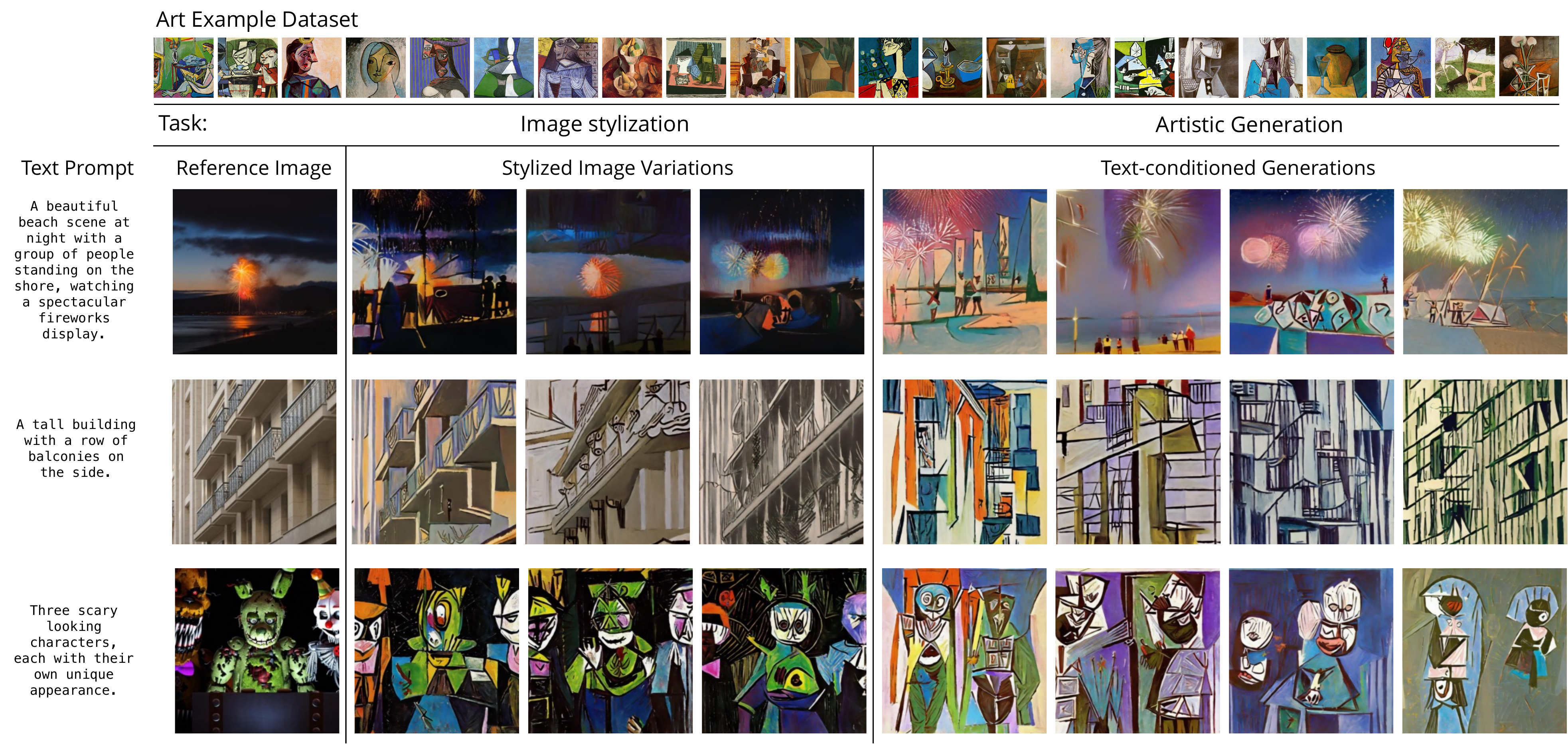}
    \caption{Additional qualitative experiments.}
    \label{fig:q:picasso}
\end{figure*}

\begin{figure*}[htbp]
    \centering
    \includegraphics[width=\linewidth]{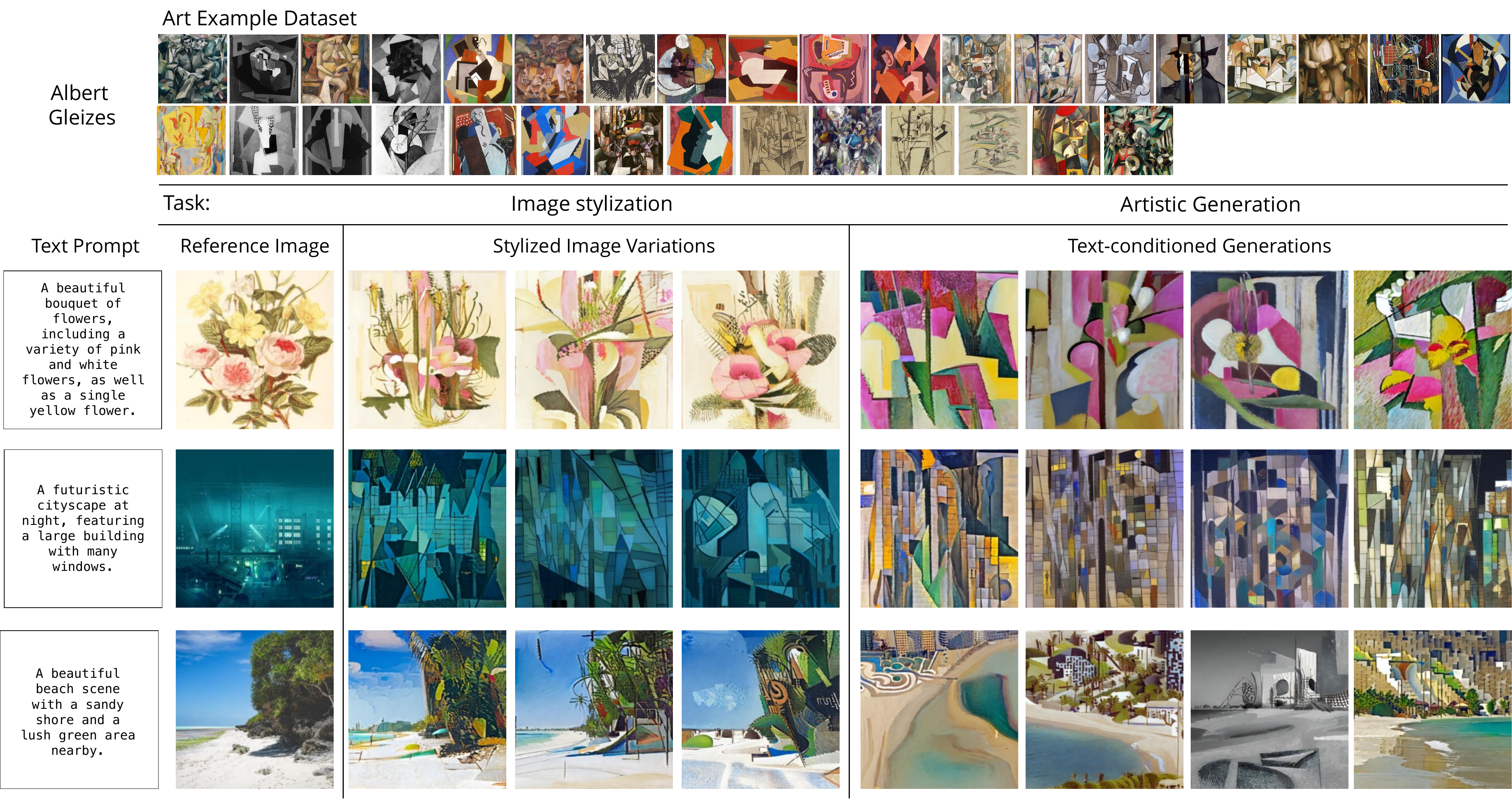}
    \caption{Additional qualitative experiments.}
    \label{fig:q:gleizes}
\end{figure*}

\begin{figure*}[htbp]
    \centering
    \includegraphics[width=\linewidth]{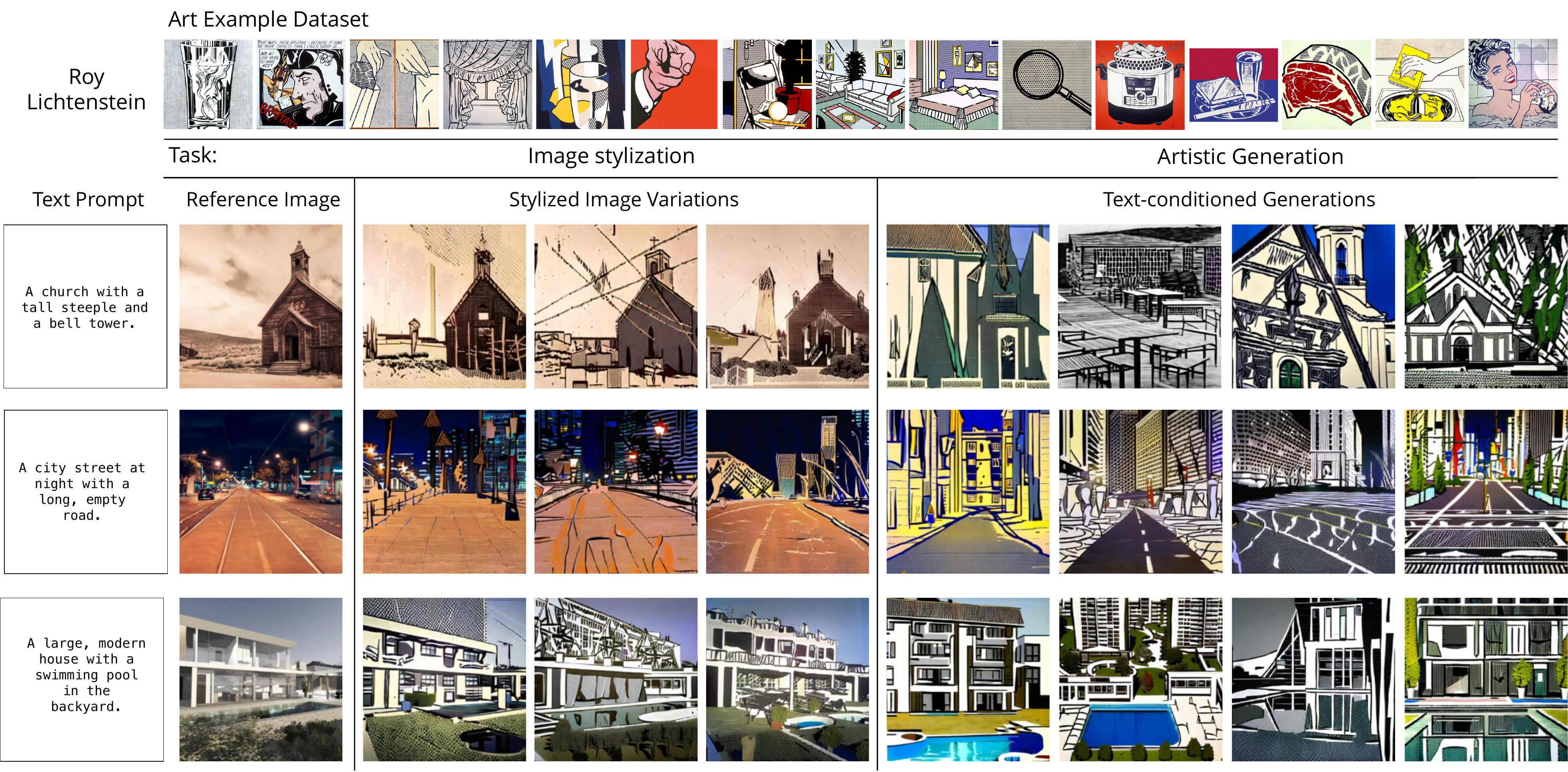}
    \caption{Additional qualitative experiments.}
    \label{fig:q:lichtenstein}
\end{figure*}

\begin{figure*}[htbp]
    \centering
    \includegraphics[width=\linewidth]{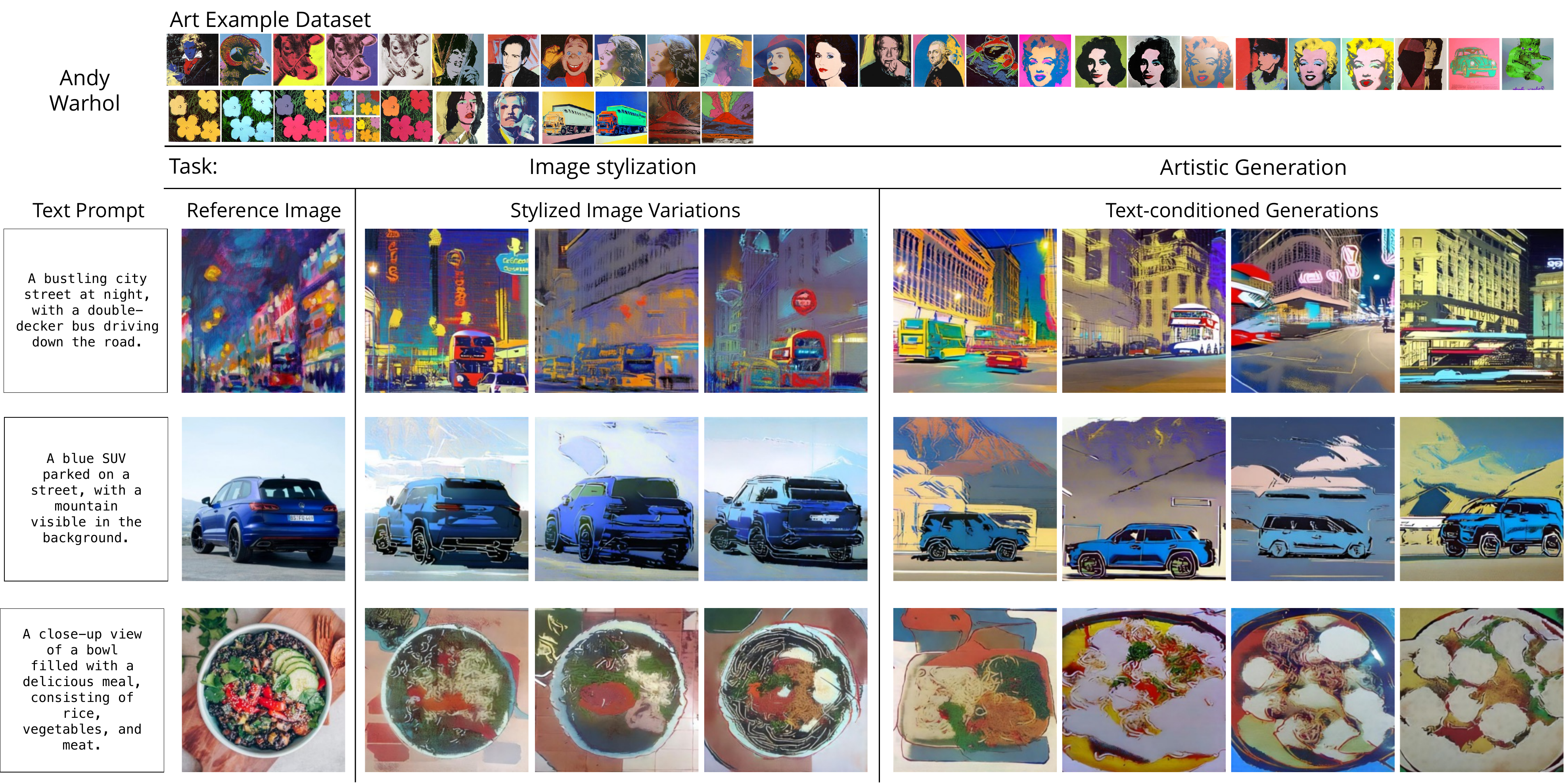}
    \caption{Additional qualitative experiments.}
    \label{fig:q:warhol}
\end{figure*}

\begin{figure*}[htbp]
    \centering
    \includegraphics[width=\linewidth]{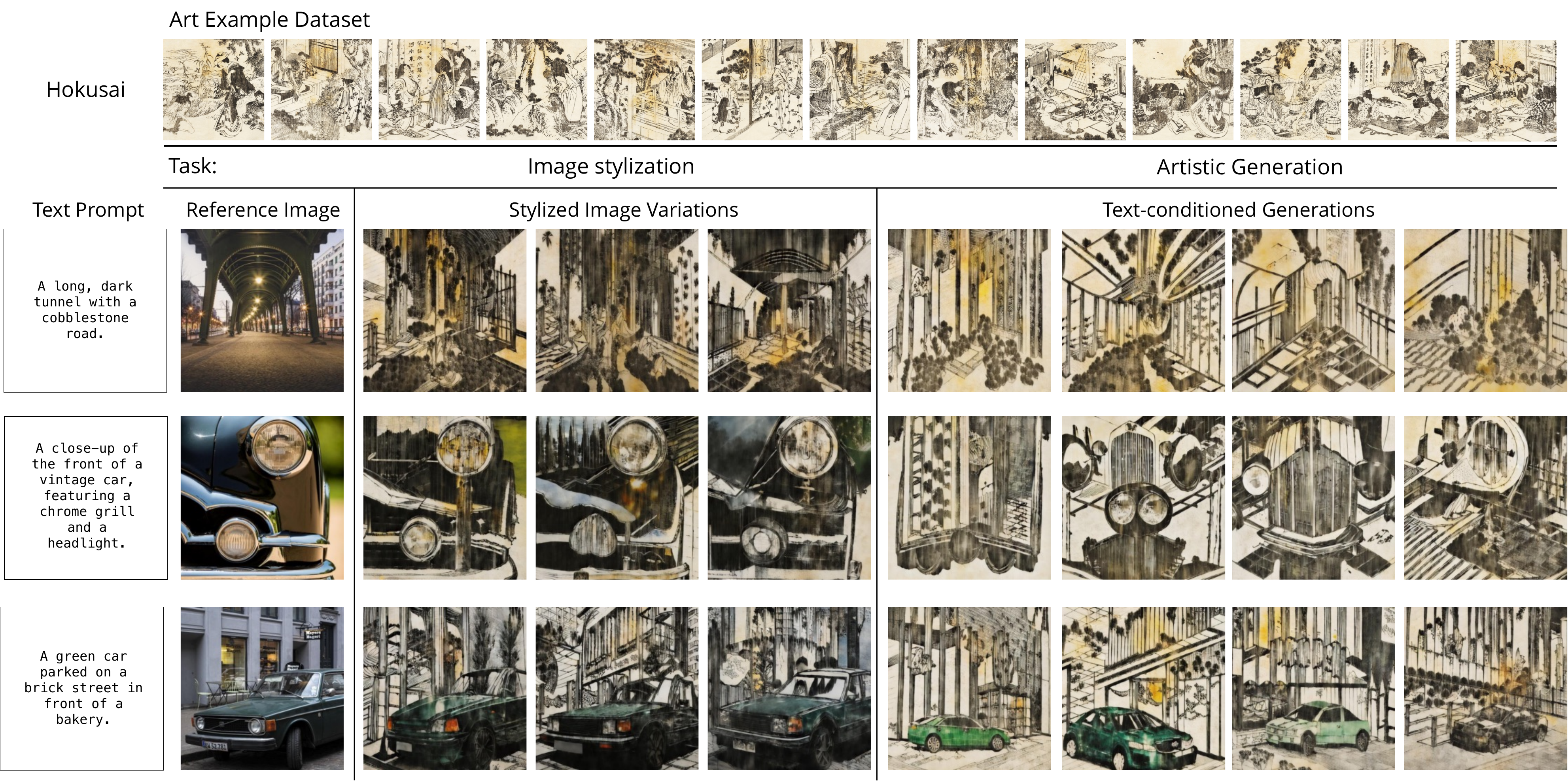}
    \caption{Additional qualitative experiments.}
    \label{fig:q:hokusai}
\end{figure*}

\begin{figure*}[htbp]
    \centering
    \includegraphics[width=\linewidth]{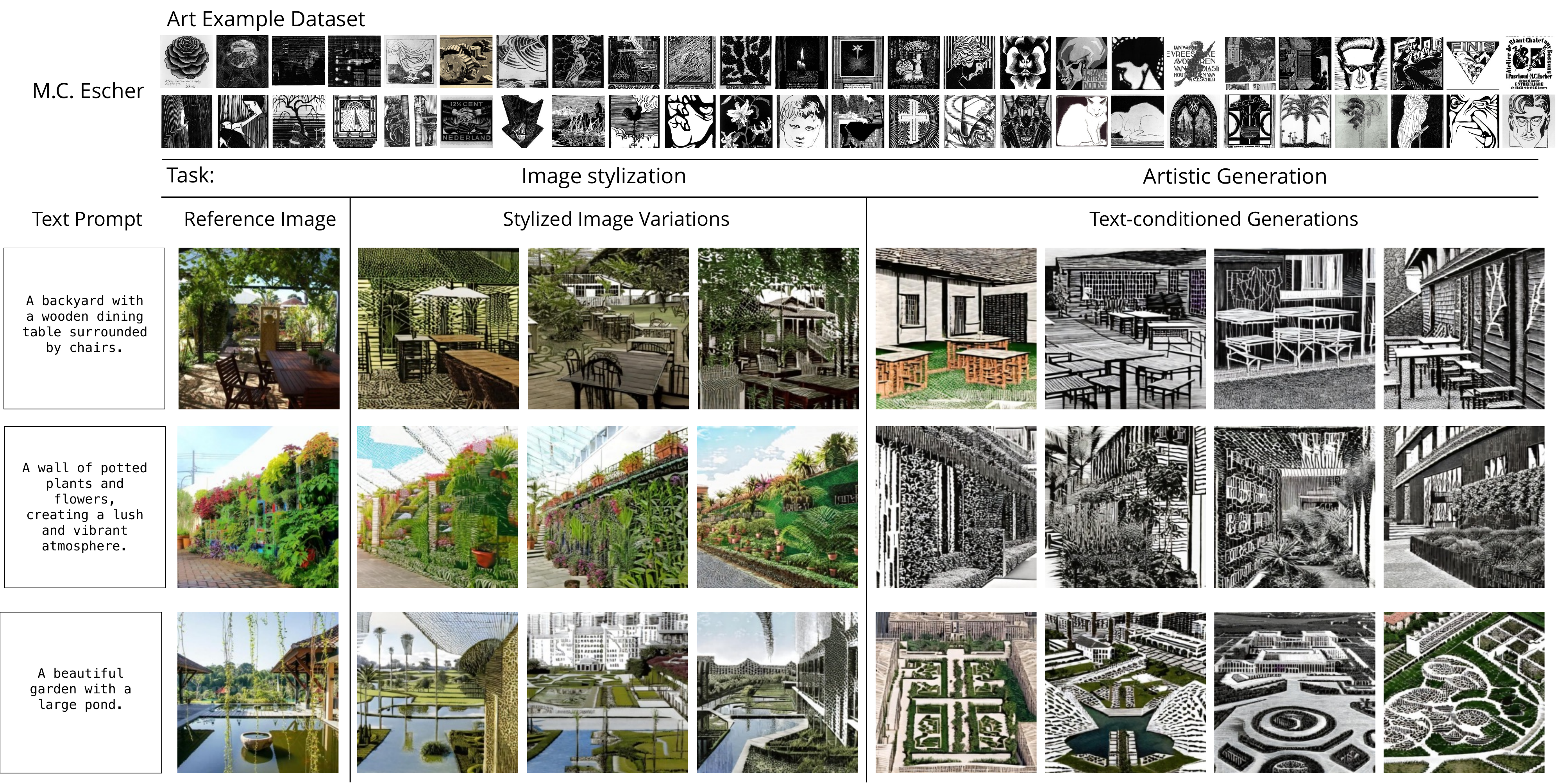}
    \caption{Additional qualitative experiments.}
    \label{fig:q:escher}
\end{figure*}

\begin{figure*}[htbp]
    \centering
    \includegraphics[width=\linewidth]{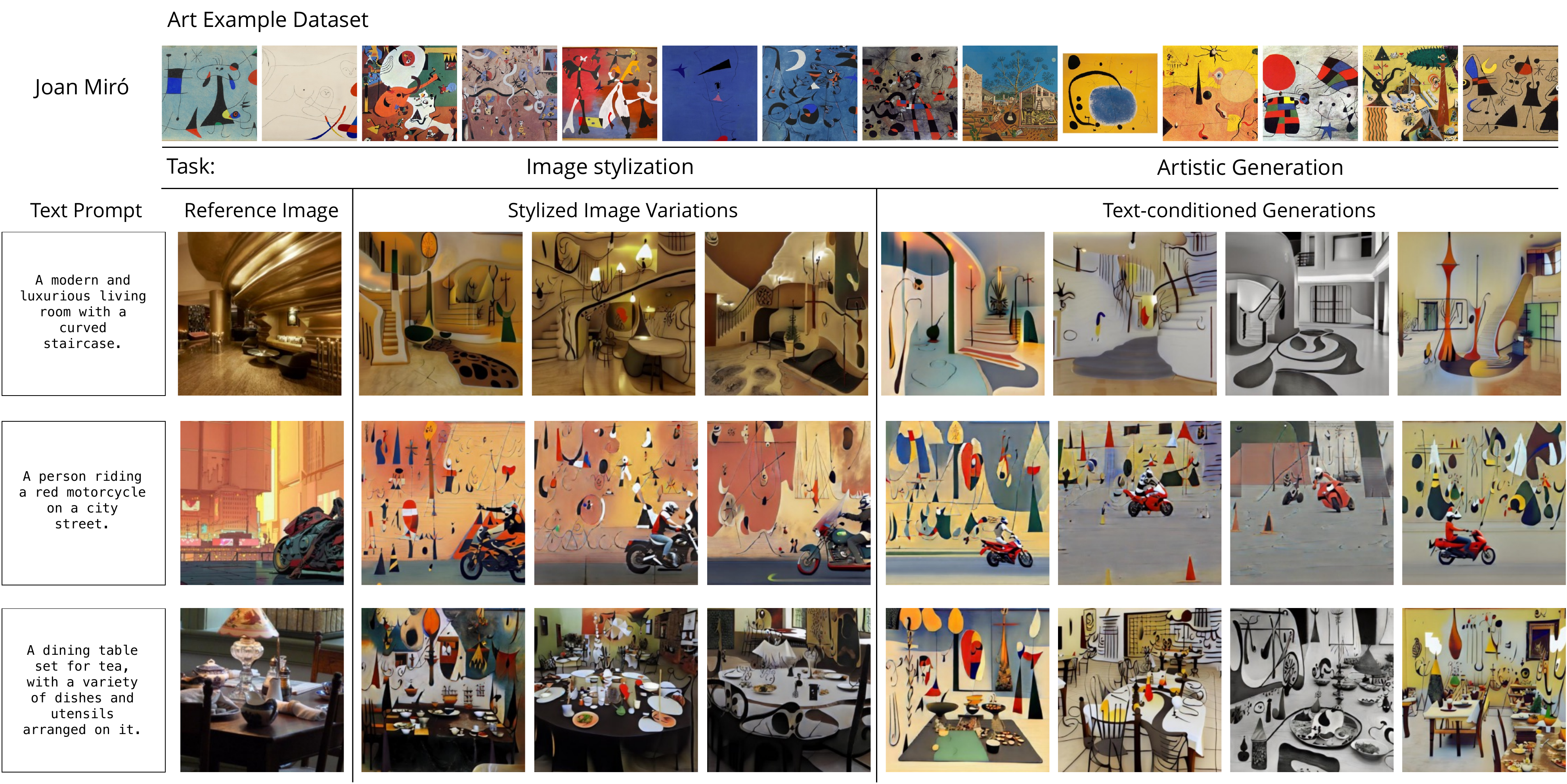}
    \caption{Additional qualitative experiments.}
    \label{fig:q:miro}
\end{figure*}

\begin{figure*}[htbp]
    \centering
    \includegraphics[width=\linewidth]{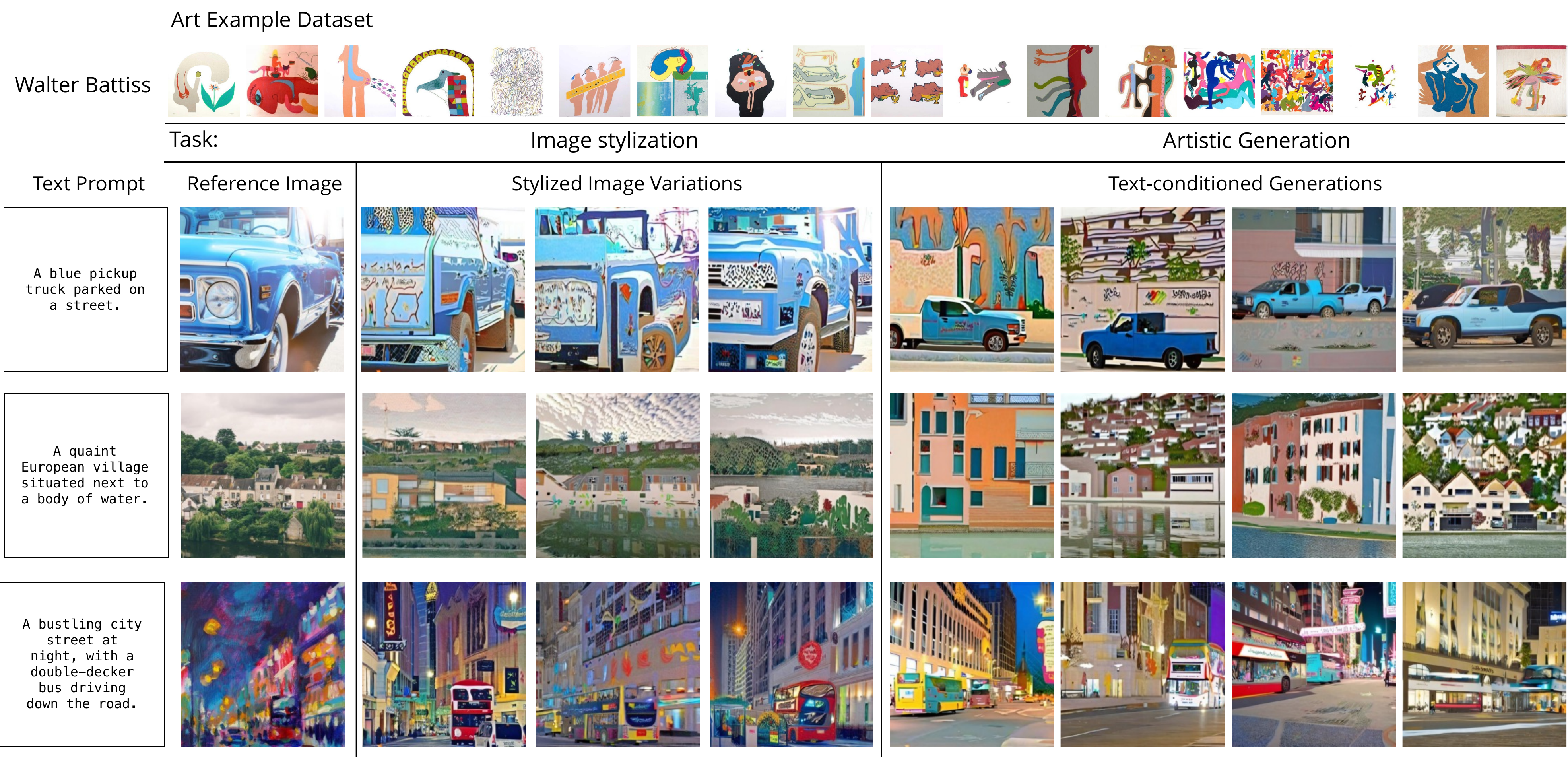}
    \caption{Additional qualitative experiments.}
    \label{fig:q:batiss}
\end{figure*}

Additional results of art generation (art generation and image stylization) and training images in Figures~\ref{fig:q:monet} --\ref{fig:q:batiss}. We show our model's ability to replicate diverse artistic styles: Impressionism (Monet, van Gogh, Corot), Art Nouveau (Klimt), Fauvism (Derain), Abstract Expressionism (Matisse, Pollock, Richter), Abstract Art (Kandinsky), Cubism (Picasso, Gleizes), Pop Art (Lichtenstein, Warhol), Ukiyo-e (Hokusai), Expressionism (Escher), and Postmodern and Geometric Abstraction (Miró, Battiss). The captions and reference images are sampled from the LAION Pop dataset.


\end{document}